\documentclass{article}

\usepackage[preprint]{neurips_2026}

\usepackage[utf8]{inputenc} % allow utf-8 input
\usepackage[T1]{fontenc}    % use 8-bit T1 fonts
\usepackage{hyperref}       % hyperlinks
\usepackage{url}            % simple URL typesetting
\usepackage{booktabs}       % professional-quality tables
\usepackage{amsfonts}       % blackboard math symbols
\usepackage{nicefrac}       % compact symbols for 1/2, etc.
\usepackage{microtype}      % microtypography
\usepackage{xcolor}         % colors

\usepackage[most]{tcolorbox}  
\usepackage[dvipsnames]{xcolor}
\usepackage{enumitem}
\usepackage{xspace}
\usepackage{caption}

\usepackage{array}
\usepackage{multirow}

\usepackage{pgfplots}
\pgfplotsset{compat=1.18} 
\usepgfplotslibrary{groupplots}
\usetikzlibrary{calc}
\usepackage{wrapfig}

\newcommand{\ourbenchmark}{{\textsc{SMFR}}\xspace} %name for the framework

\title{The Illusion of Multi-Agent Advantage}

\author{%
  Prathyusha Jwalapuram\thanks{Equal contribution.
  $^{\dagger}$Project lead.
  $^{\ddagger}$Project advisor.
  $^1$Salesforce Research.
  $^2$HKUST (Guangzhou).
  $^3$University of British Columbia.
  $^4$Nanyang Technological University.
  Correspondence to: Prathyusha Jwalapuram <pjwalapuram@salesforce.com>, Hehai Lin <hlin709@connect.hkust-gz.edu.cn>, 
  Zixuan Ke <zixuan.ke@salesforce.com>, and Shafiq Joty <sjoty@salesforce.com>.}\space\space{$^{1}$}
  \\
  \And
  Hehai Lin $^{*2}$\\
  \And
  Chuyuan Li $^{3}$\\
  \And
  Fangkai Jiao $^{4}$\\
  \And
  Sudong Wang $^{2}$\\
  \And
  Yifei Ming $^{1}$\\
  \And
  Zixuan Ke $^{\dagger 1}$\\
  \And
  Chengwei Qin $^{2}$\\
  \And
  Giuseppe Carenini $^{3}$\\
  \And
  Shafiq Joty $^{\ddagger 1}$\\
}

\begin{document}

\maketitle

\begin{abstract}
Prevailing wisdom posits that Multi-Agent Systems (MAS) are superior to Single-Agent Systems (SAS), citing advantages like {context protection,} parallel processing and distributed decision-making. However, empirical support for this claim relies primarily on comparisons with SAS baselines using benchmarks that prioritize isolated reasoning tasks, which do not adequately assess these advantages. Focusing on automatically generated MAS that are designed for enhanced generalizability over manually-designed counterparts, we perform a rigorous, systematic evaluation against SAS, specifically Chain-of-Thought with Self-Consistency (CoT-SC). Across traditional reasoning datasets and tasks with interactive multi-step workflows (\textit{e.g.,} BrowseComp-Plus), we demonstrate that automatic MAS consistently underperform CoT-SC despite being up to 10x more expensive. To isolate these failures from limitations inherent to task structure, we introduce a diagnostic synthetic dataset tailored for MAS featuring explicit task decomposition, context separation and parallelization potential. We show that expert-architected MAS consistently outperforms automatically generated architectures in both raw performance and cost-efficiency on this dataset, demonstrating that existing evaluation frameworks mask critical architectural gaps and inefficiencies of complex MAS by failing to account for the marginal utility of increased computational cost. Critically, systematic deconstruction of the generated MAS architectures reveals that current automated design paradigms produce architectural bloat that prioritizes superficial complexity which does not translate into functional utility, exposing a fundamental misalignment with multi-agent principles.\footnote{The dataset and code can be found at \url{https://multi-agent-eval.github.io/}.}

\end{abstract}

\section{Introduction}
\label{sec:intro}
\begin{figure}
\centering
% \vspace{-2\baselineskip}
\includegraphics[width=0.9\columnwidth]{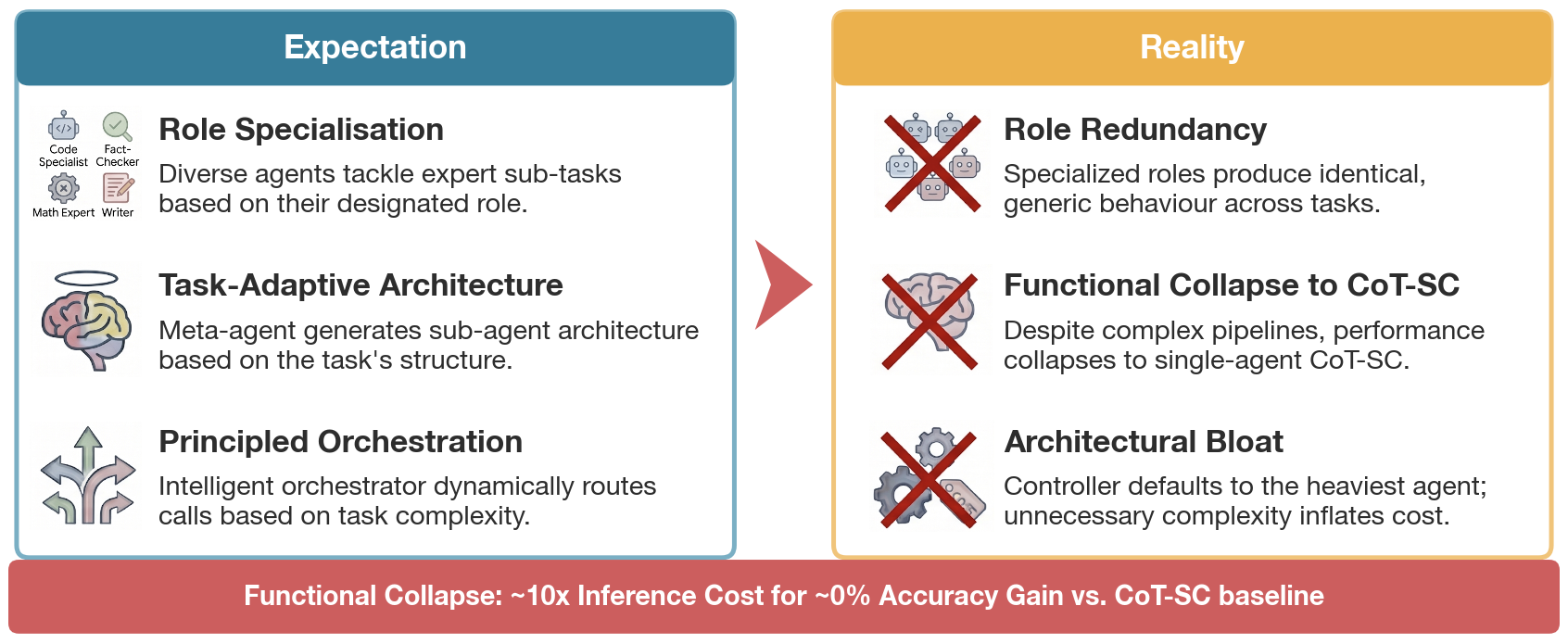}
\vspace{-0.5\baselineskip}
\captionof{figure}{\textbf{The Illusion of Multi-Agent Advantage.} Theory promises specialization (left); reality reveals redundancy and functional collapse (right). Automated frameworks often incur $\approx 10\times$ the cost of CoT-SC for negligible gains (Section~\ref{sec:analysis}).}

\label{fig:teaser}
\vspace{-1\baselineskip}
\end{figure}

Although Large Language Models (LLMs) have evolved significantly in their capabilities, they alone as Single-Agent Systems (SAS) still fall short on several complex reasoning tasks, such as BrowseComp-Plus \citep{chen2025BrowseCompPlus} and Humanity's Last Exam (HLE) \citep{phan2025humanity}. Multi-Agent Systems (MAS) have increasingly been introduced as a solution \cite{ke2025surveyfrontiersllmreasoning}, under the assumption that multiple coordinated LLM agents would outperform SAS by enabling collective decision making through mechanisms such as task decomposition, parallel execution, context separation, role specialization, debate, reconciliation, and cross-verification \citep{Foerster2016LearningTC, HernandezLeal2018ASA, Wang2019AchievingCT, Wang2024MixtureofAgentsEL, Zhou2025MultiAgentDO, Gao2025SingleagentOM}. 

This expectation has led to the rapid development of \textbf{automatic MAS}, characterized by an automated coordination layer that dynamically decomposes tasks, configures agent roles, routes information, and manages execution flow \citep{Ke2025MASZero, hu2024ADAS, zhang2024aflowautomatingagenticworkflow, liu2024dynamicllmpowered, zhang2025agentic-supernet, Ke2026MASOrchestra}, in contrast to manually-designed MAS, which rely on substantial human effort and often lack generalizability to novel tasks. Automatic MAS can also be designed as decentralized agent-team systems, where agents communicate and act more independently \citep{anthropic2026claudecodeagentteams,openclawagents2026,mirofish2026}. While decentralized systems offer alternative scaling properties, centralized coordination  paradigms represent the current standard for high-precision task execution and are thus the focus of our evaluation.

Despite their popularity, the realized advantages of automatic MAS remain unclear. Most evaluations compare MAS against SAS baselines such as Chain-of-Thought (CoT) \citep{wei2022chain}, CoT with Self-Consistency (CoT-SC) \citep{wang2023selfconsistency}, or self-refinement \citep{madaan2023selfrefine}, reporting improved accuracy in tasks such as mathematical reasoning \citep{Chen2023AgentVerseFM, Zhou2025MultiAgentDO, Ke2025MASZero, hu2024ADAS, dylan, zhang2024aflowautomatingagenticworkflow, zhang2025agentic-supernet}, question answering (QA) \citep{Zhou2025MultiAgentDO, Ke2025MASZero, hu2024ADAS, dylan, zhang2024aflowautomatingagenticworkflow}, and coding \citep{Zhang2025MetaAgentAC, Chen2023AgentVerseFM, Zhou2025MultiAgentDO, Ke2025MASZero, dylan, zhang2024aflowautomatingagenticworkflow, zhang2025agentic-supernet}. However, these comparisons rarely control for inference budgets such as number of LLM calls, total cost, retries, or sampled paths. Thus, MAS may appear stronger due to increased test-time computation rather than superior coordination \citep{kapoor2024ai, anthropic2025multiagentresearch}. Recent studies further question MAS robustness, showing inconsistent performance against strong SAS baselines \citep{Gao2025SingleagentOM} and instability in debate or verification mechanisms \citep{Wynn2025TalkIA, venkataramani2026masproveunderstandingprocessverification}. While controlled analyses suggest gains depend on task topology and cost \citep{Kim2025TowardsAS, Ke2026MASOrchestra}, existing evaluations focus on outcome accuracy rather than whether motivating mechanisms - such as task decomposition, parallelization, or context separation - actually manifest in automated workflows.

Moreover, Anthropic \cite{anthropic2025whentomultiagentresearch} recommends building MAS for tasks where context separation provides clear benefits like a) protection of context, b) parallelization and c) specialization in terms of domain, system prompt, tool set, etc. Although recent datasets such as MASBench \citep{Ke2026MASOrchestra} provide a controlled framework for analyzing MAS behavior, most evaluations still rely on tasks originally designed for SAS, which do not isolate properties such as sub-task structure, parallel execution, or role specialization. \cite{Kim2025TowardsAS} categorize commonly used reasoning and QA benchmarks such as GSM8K \cite{cobbe2021gsm8k} and MMLU \cite{hendrycks2021measuring} as unsuitable for evaluating agentic capabilities, since they evaluate static reasoning, contrasting them against benchmarks such as BrowseComp-Plus \cite{chen2025BrowseCompPlus}, which requires dynamic and progressive information seeking and reasoning. To address this apparent misalignment in current MAS evaluation paradigms, our work centers on three primary investigations:
\begin{center}
\begin{enumerate}
[leftmargin=*,noitemsep,topsep=1pt]
    \item \textbf{Comparative Performance:} Do automatic MAS provide consistent and cost-effective performance advantages over strong SAS baselines?
    \item \textbf{Isolating Task Suitability:} When provided with explicit structural opportunities for multi-agent execution, can automated orchestrators translate these into functional utility?
    \item \textbf{Architectural Alignment:} Do automated systems successfully manifest core MAS principles like parallelization, specialization and context protection?
\end{enumerate}
\end{center}

We measure \textbf{comparative performance} by conducting systematic evaluations of automatic MAS against strong SAS baselines, particularly CoT-SC. Our evaluation spans multiple model sizes and families, including GPT-4o, GPT-5, GPT-OSS-120B, and Gemini-2.5-Pro, and covers both standard reasoning tasks and more complex agentic settings such as GPQA-Diamond, HLE-Maths, SWE-Bench Lite, and BrowseComp-Plus. We find that automatic MAS do not consistently outperform SAS; in many settings, CoT-SC matches or exceeds MAS performance while being more cost-efficient.

To \textbf{isolate task suitability} as a contributing factor and to investigate if MAS principles emerge under favorable conditions, we introduce the \textbf{Synthetic Multi-Hop Financial Reasoning} (\ourbenchmark) dataset. \ourbenchmark features an explicit sub-task structure, context-heavy inputs, and clear opportunities for parallelization and specialization. We find once again that CoT-SC reliably outperforms automatic MAS on this task, demonstrating that task suitability is not a factor in their poor performance. We also construct an expert-designed MAS baseline with explicit decomposition, role specialization, and deterministic orchestration. This baseline performs strongly, demonstrating that tasks can benefit from MAS when the system is properly structured. 

We further analyze the \textbf{architectural alignment} of the generated MAS with core MAS principles. Our deconstruction of these workflows shows \textbf{architectural bloat and systematic failure in core agentic functions}. Specifically: (i) assigned agent roles are often functionally redundant; (ii) many automated MAS effectively collapse into basic CoT-SC execution; and crucially (iii) this lack of specialization is consistent across disparate tasks, exposing a fundamental deficit in adaptive task decomposition. Together, our findings suggest that the perceived advantage of automated MAS is often a byproduct of superficial complexity rather than structural synergy. Our contributions include:
\begin{enumerate}
[leftmargin=*,noitemsep,topsep=2pt]
    \item \textbf{A Critical Re-evaluation of the MAS Advantage:} We demonstrate through systematic benchmarking that automated MAS rarely outperform SAS baselines when accounting for cost-efficiency and baseline strength.
    \item \textbf{The \ourbenchmark{} Diagnostic Benchmark}: We introduce \textbf{Synthetic Multi-Hop Financial Reasoning}, a diagnostic task featuring explicit sub-structures and a gold-standard \textbf{Expert-MAS} to establish an empirical performance upper bound for MAS.
    \item \textbf{Architectural Deconstruction:} We provide a rigorous analysis of synthesized MAS workflows, exposing functional collapse where complex automated designs revert to basic single-agent execution in practice.

\end{enumerate}

\section{Related Work}
While ``agenticness'' exists on a continuum \citep{kapoor2024ai}, we distinguish Single-Agent Systems (SAS) from Multi-Agent Systems (MAS) based on the locus of reasoning. We define SAS as a single sequential control loop governed by one LLM instance, encompassing tool use \citep{yao2023react}, self-reflection \citep{madaan2024self}, and CoT reasoning. In contrast, MAS features multiple LLM-backed agents interacting through structured protocols \citep{xi2023risepotentiallargelanguage}, where behavior emerges from collective reasoning. Our work specifically evaluates centralized, automated MAS, where an orchestrator dynamically manages roles and information flow, as these frameworks represent the current frontier of agentic scaling.

\textbf{Inference-time Automatic MAS.} These MAS adapt the agent configuration dynamically for each query. DyLAN \citep{dylan} utilizes importance scoring to select sub-agents on the fly, while MAS-Zero \citep{Ke2025MASZero} attempts zero-shot coordination without external validation. 

\textbf{Optimized Automatic MAS.} To minimize test-time overhead, these frameworks discover or train optimal architectures prior to deployment. ADAS \citep{hu2025automated} and AFlow \citep{zhang2024aflowautomatingagenticworkflow} treat MAS design as a code-generation task, utilizing Monte Carlo Tree Search (MCTS) to find workflows that perform well on a validation set. Others, such as ToolOrchestra \citep{su2025toolorchestraelevatingintelligenceefficient} and MAS-Orchestra \citep{Ke2026MASOrchestra}, use Reinforcement Learning (RL) to train a centralized orchestrator. Frameworks like MaAS \citep{zhang2025agentic-supernet} occupy a middle ground; while the underlying operator distributions are pre-optimized, the system performs inference-time routing by sampling query-dependent architectures on the fly. 

We evaluate both kinds of systems to determine if dynamic flexibility or pre-optimized workflows justify the significant per-query compute overhead without architectural bloat and functional collapse.

\textbf{Diagnostics of Multi-Agent Failure.} \cite{cemri2025multiagentllmsystemsfail} categorize execution-level failures (\textit{e.g.,} communication lapses), whereas we diagnose structural inefficiencies (\textit{e.g.,} role redundancy, functional collapse), inherent to automated MAS search. While \cite{kapoor2024ai} and \cite{Kim2025TowardsAS} question benchmark suitability for MAS, we introduce \ourbenchmark as a diagnostic tool to isolate task suitability. Finally, addressing \cite{tran2026singleagentllmsoutperformmultiagent}'s critique regarding compute-confounded gains, we show that CoT-SC consistently outperforms MAS despite a significantly lower token budget. This indicates that current automated designs suffer from architectural bloat, failing to translate high expenditure into reasoning gains.

\section{Critical Re-Evaluation of the MAS Advantage }
\label{sec:experiments}

To investigate whether automatic MAS show consistent and cost-effective performance advantages over strong SAS baselines, we conduct a large-scale audit comparing SAS and MAS performance and cost across multiple LLM model sizes and families, covering standard reasoning tasks and complex agentic settings. We specifically test the hypothesis that MAS provide a superior scaling path compared to simple, budget-matched ensembling.

\subsection{Experimental Setup}
\label{subsec:experiment_setup}

\textbf{Benchmark Datasets.} As detailed in Section~\ref{sec:intro}, mathematical reasoning, QA, and coding are the primary domains for evaluating MAS. Following standard practice, we select the most up-to-date and challenging variants of these tasks to attempt a systematic reproduction of commonly reported MAS improvements, ensuring our evaluation reflects the current performance ceiling of the field. Specifically, we target: (i) mathematical reasoning through \textbf{HLE-Maths} \cite{phan2025humanity}; (ii) QA through \textbf{GPQA-Diamond} \cite{rein2023gpqagraduatelevelgoogleproofqa}; and (iii) code generation through \textbf{SWE-Bench Lite} \cite{swe-bench}. However, since these benchmarks prioritize static reasoning, we also include \textbf{BrowseComp-Plus} \cite{chen2025BrowseCompPlus} following the recommendation from \cite{Kim2025TowardsAS} to provide a critical test bed for progressive information seeking and dynamic reasoning. By utilizing these state-of-the-art variants, we aim to reproduce commonly reported MAS improvements and assess their robustness under stringent conditions.

\textbf{Automatic MAS Baselines.} We select six representative frameworks that span the current state-of-the-art in autonomous agent coordination, including both inference-time and optimized (training/validation based) variants (see Appendix~\ref{app:model_config} for complete experimental setup configuration details): 

\begin{itemize}[leftmargin=*,noitemsep,topsep=2pt]
    \item \textbf{DyLAN} \citep{dylan} iteratively selects top-$K$ specialized agents via LLM-ranking, using dynamic interaction layers to refine team composition from a diverse pool of roles.
    
    \item \textbf{MAS-Zero} \citep{Ke2025MASZero} is a zero-shot framework where a meta-agent iteratively optimizes multi-agent orchestrations by selecting from four reasoning blocks (CoT, CoT-SC, Reflexion, and Debate). A verifier then evaluates all generated candidate trajectories to select the final response.
    
    \item \textbf{ADAS} \citep{hu2025automated} employs a meta-agent to iteratively discover agentic architectures by generating novel coordination code. Performance metrics from these implementations are stored in an archive to guide subsequent discovery iterations via validation data.
    
    \item \textbf{AFlow} \citep{zhang2024aflowautomatingagenticworkflow} treats workflow design as code-based search, utilizing Monte Carlo Tree Search (MCTS) with an LLM-based optimizer to iteratively refine candidates based on validation feedback.

    \item \textbf{MaAS} \citep{zhang2025multiagentarchitecturesearchagentic} uses a controller to sample query-dependent workflows from a probabilistic supernet, sequentially activating operators until a threshold is met. This architecture facilitates dynamic early exits and is optimized via textual gradients from environmental feedback.
    
    \item \textbf{MAS-Orchestra} \citep{Ke2026MASOrchestra}
    employs an RL-trained orchestrator to manage sub-agent delegation. System complexity is governed by the Degree of MAS (DoM), where the orchestrator selects sub-agent configurations (e.g., CoT, Debate) from a fixed candidate pool based on task requirements.
 
\end{itemize}

\textbf{Backbone LLMs.} To ensure generalization across paradigms, we evaluate with a stratified selection of LLMs: \textbf{GPT-4o}, \textbf{GPT-5}, \textbf{GPT-OSS (120B)}, and \textbf{Gemini-2.5-Pro}. This ensemble spans frontier closed-source models, varied generations, and open-source alternatives. While resource and API cost considerations necessitated a focused set of backbone models, this cross-section allows us to determine if architectural gaps are systemic across different model families and scales.

\begin{figure}[t]
    \centering
    \includegraphics[width=\linewidth]{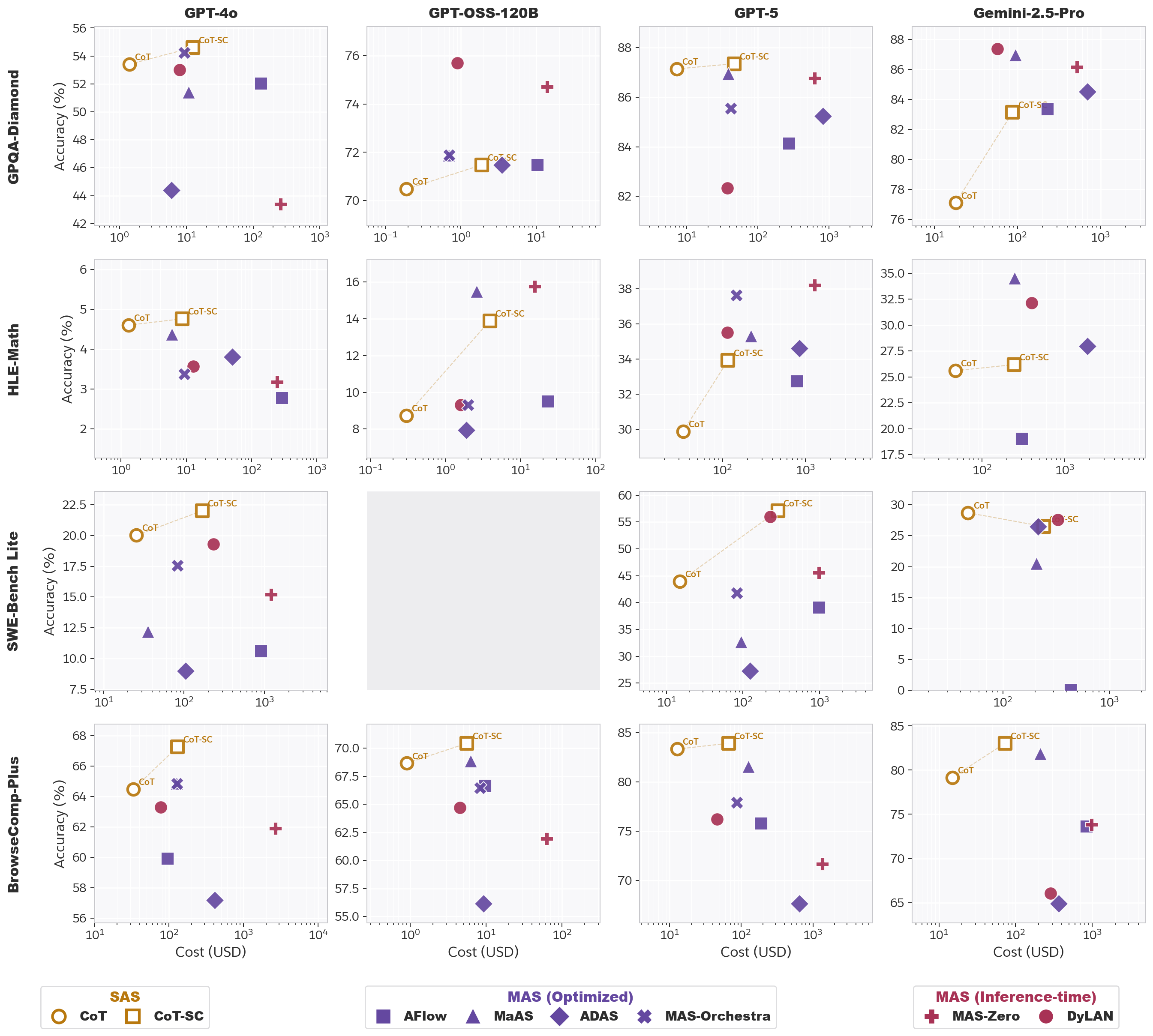}
    \caption{\textbf{The MAS Efficiency Frontier.} Cost vs. accuracy trade-offs. \textbf{CoT-SC} provides the optimal balance of performance and cost-efficiency. Automated MAS (e.g., \textbf{ADAS}, \textbf{MAS-Orchestra}) frequently incur $10\times$ inference costs vs. SAS baselines for negligible gains, except on \textbf{HLE-Math}. This suggests MAS fails to elevate weaker backbones. Note: GPT-OSS-120B was excluded from \textbf{SWE-Bench Lite} due to consistent formatting failures in code patches.}
    \label{fig:cost_vs_perf_all_models}
\end{figure}

\textbf{Evaluation Protocol.} Results are averaged across 3 independent runs.\footnote{Gemini-2.5-Pro results use a single run due to cost.} CoT-SC baseline employs a 5-sample majority vote across all datasets and backbones. Appendix~\ref{sec:app_datasets} details test splits. 

\subsection{Results}
\label{subsec:results}

Figure~\ref{fig:cost_vs_perf_all_models} visualizes the cost-benefit profile of MAS performance relative to inference expenditure (including search and validation overhead; see Table~\ref{tab:full_raw_results} for the full results). A dominant trend emerges across all benchmarks: CoT-SC consistently outperforms automated MAS frameworks, frequently achieving higher accuracy at less than $10\%$ of the computational cost. This suggests that for current frameworks, architectural complexity is an inefficient substitute for simple stochastic sampling.

\textbf{Challenging Capability Bridging.} Our results directly challenge the prevailing assumption that sophisticated orchestration can elevate weaker models to frontier-level performance \citep{li2024more}: (i) \textbf{No Gains for Mid-Tier Models:} MAS fails to provide consistent improvements for models like GPT-4o or GPT-OSS; (ii) \textbf{Model Tier Superiority:} A single-agent GPT-5 instance using CoT-SC reliably outperforms the most sophisticated GPT-4o-based MAS frameworks (e.g., ADAS or AFlow) while consuming less than half the total tokens. These findings indicate that automated MAS designs cannot bridge the generational gap between model tiers; instead, they introduce significant computational bloat without commensurate gains.

\textbf{Complexity Requires Competence.} Interestingly, significant MAS uplift only occurs on HLE-Math using GPT-5 and Gemini-2.5-Pro. This suggests a competency floor for MAS: architectural complexity may only yield benefits when the underlying backbone already possesses the high inherent reasoning capabilities necessary to navigate complex coordination.

\textbf{Takeaways.} Overall, these findings provide empirical evidence for architectural bloat across the MAS ecosystem. The significant performance-cost gap suggests that the sophisticated multi-agent graphs generated by these frameworks do not translate into functional reasoning gains. Instead, they represent a failure of automated search to find configurations that outperform unstructured scaling, confirming that current MAS designs have yet to move beyond redundant high-cost iterations.

\subsection{The \ourbenchmark Diagnostic Benchmark}
\label{sec:smfr_benchmark}

Results from Section~\ref{subsec:results} show that CoT-SC outperforms MAS across all standard benchmarking datasets in both accuracy and cost-effectiveness. However, existing works such as \cite{kapoor2024ai, Kim2025TowardsAS} have critiqued the use of benchmarks created under the assumption of simple input-output flows for testing MAS. To isolate task suitability as a factor for the poor performance of MAS, we create a task tailored for multi-agent workflows called the \textbf{Synthetic Multi-Hop Financial Reasoning (\ourbenchmark)} dataset. 

\textbf{Task Structure.} Each problem presents an agent with a stock price haystack - historical open/close prices for $B$ companies over a 30-day window - and a set of investor transactions (buy/sell pairs). The agent must determine on which dates each investor could achieve a specified profit or loss target, then identify the winning investor according to an aggregation criterion (earliest or latest qualifying date). The task is designed to resist shortcut strategies: correct answers require multi-step context extraction (price lookup, date lookup, date filtering) and numerical reasoning (P\&L computation, target price derivation, sorting). Figure~\ref{fig:smfr_sample} shows an example instance.

\begin{figure}
    \centering
    \includegraphics[width=\linewidth]{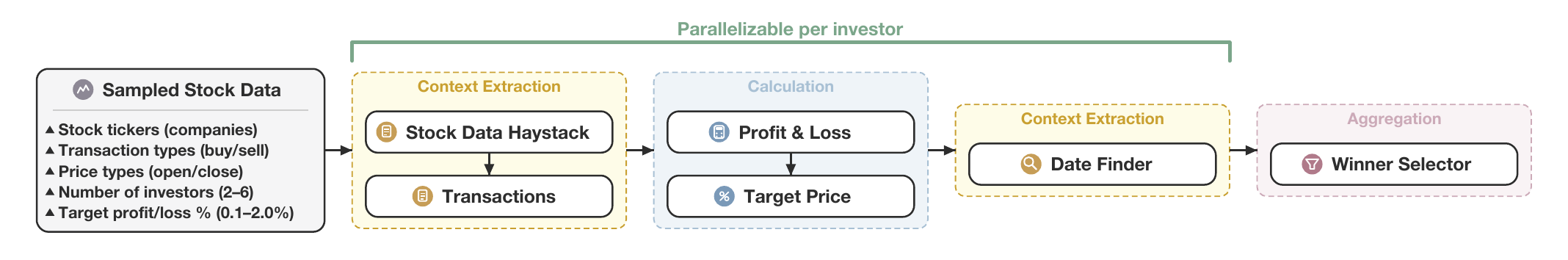}
    \caption{\textbf{SMFR Dataset Generation Pipeline.} Stock data from \cite{yfinance2024} is sampled along with parameters such as transaction type, price type, number of investors, etc. Price tables with distractor data are used to create a haystack; specific transaction prices and dates for investors are the needles that need to be retrieved. The P\&L calculations and winning investor (answer) is programmatically computed.} %\shafiq{Needs better organization of the textboxes..}\zixuan{may refer to fig. 3 in https://arxiv.org/pdf/2505.06120}}
    \label{fig:stock_data_gen}
\end{figure}

\textbf{Non-Linear Interdependence.} \cite{zhu2025establishing} establish guidelines for creating agentic benchmarks, which include requirements such as sequential interdependence, where later actions must depend on earlier observations. Anthropic \cite{anthropic2025whentomultiagentresearch} recommends that MAS are suitable for tasks where context separation, task parallelization and specialization provide clear benefits. Following their recommendations, \ourbenchmark is explicitly designed to be \textit{non-linear} and \textit{context-heavy}. Unlike standard QA or mathematical tasks, \ourbenchmark cannot be solved via greedy local reasoning. It requires maintaining a global objective (target profit) while executing independent, modular sub-tasks (investor-specific P\&L), including retrieval of information from a large context of historical market data. A correct solution requires (i) \textbf{Constraint Parsing} (defining targets and comparison logic); (ii) \textbf{Transaction Extraction} (parsing haystack positions); (iii) \textbf{P\&L Derivation} (establishing realized baselines); (iv) \textbf{Reverse-Price Calculation} (deriving required target prices); (v) \textbf{Threshold Scanning} (validating dates); and (vi) \textbf{Cross-Investor Synthesis} (aggregating and selecting the final answer).

Figure~\ref{fig:stock_data_gen} details the task generation pipeline. \textbf{Transaction Extraction}, \textbf{Portfolio P\&L Derivation}, and \textbf{Reverse-Price Calculation} provide explicit opportunities for paralellization across investors, while being sequentially dependent within each investor's trajectory. 

\textbf{Synthetic Data Generation.} We programmatically generate problems using historical US equity prices \citep{yfinance2024}. Each instance follows a ``Needle-in-a-Haystack'' architecture (Figure~\ref{fig:stock_data_gen}): the ``Haystack'' comprises 30-day price tables for $B$ stocks; the task requires models to retrieve specific investor histories (the ``Needles'') and compute the exact date a target profit/loss threshold was achieved for an open position. As the specific problem instances are procedurally generated, the benchmark remains immune to data contamination while maintaining the realistic price distributions essential for robust evaluation. The dataset consists of $588$ test samples ($+16$ for validation) balanced across transaction types, aggregation logic, and target percentages (more details and statistics in Appendix~\ref{app:synthetic_data_gen}).

\begin{figure}[t]
    \centering
    \includegraphics[width=\linewidth]{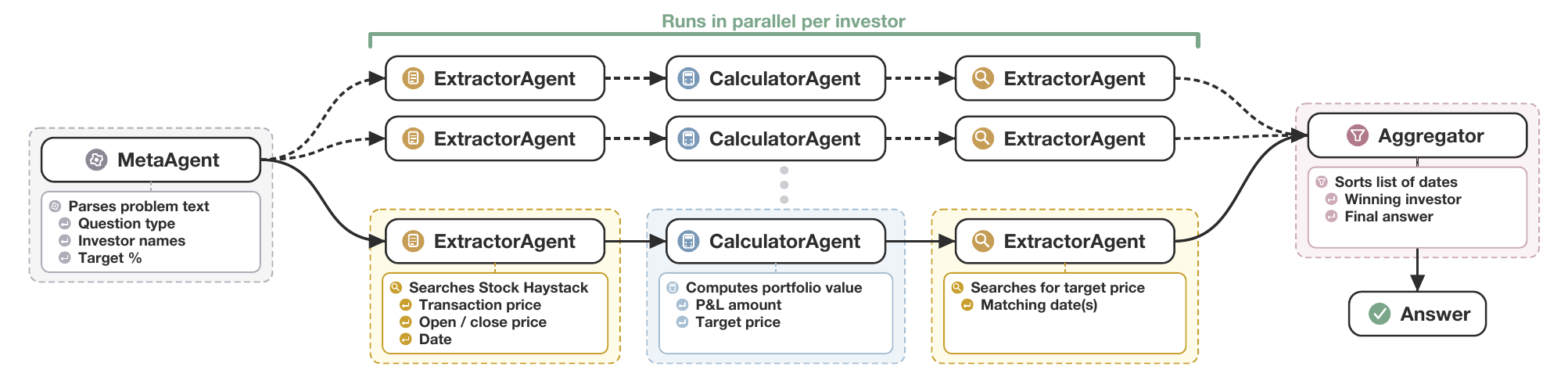}
    \caption{\textbf{Expert-MAS Pipeline Architecture.} A deterministic, code-driven architecture serving as competitive baseline. The system enforces separation of concerns: (1) Meta-Agent parses task topology, (2) ExtractorAgent retrieves targeted data, and (3) CalculatorAgent reasons over isolated snippets. A Python orchestrator dispatches these chains concurrently per investor, with final comparisons computed deterministically to ensure high-precision, low-noise consistency. }

    \label{fig:manual-mas}
\end{figure}

\textbf{Expert-Designed MAS.} To establish a competitive reference baseline for performance on \ourbenchmark, we design an \textbf{Expert-MAS} based on guidelines from Anthropic \cite{anthropic2025whentomultiagentresearch} that utilizes structured decomposition and deterministic orchestration. Expert-MAS enforces a strict separation between context processing and logical control (Figure~\ref{fig:manual-mas}), decomposing the task into a multi-step pipeline where a Meta-Agent first parses the problem topology into a structured schema. This schema then drives a deterministic Python-based Executor that orchestrates specialized sub-agents for targeted retrieval and numerical reasoning. By offloading task coordination and final win-determination to deterministic code, Expert-MAS minimizes context bloat and eliminates the ``orchestration noise'' prevalent in automated MAS designs. Appendix~\ref{app:manual_mas} details the full configuration setup. 

\begin{figure}
    \centering
    \includegraphics[width=0.85\linewidth]{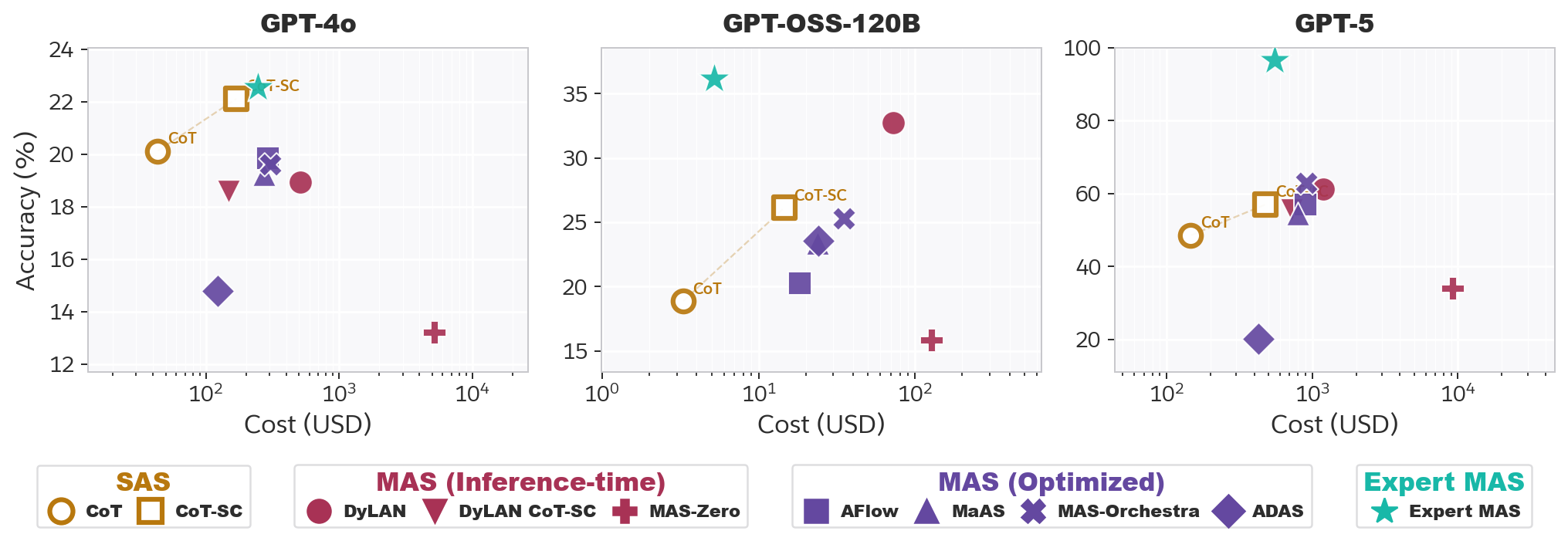}
    \caption{Automated MAS consistently fail to surpass \textbf{CoT-SC} efficiency on \ourbenchmark as well. Expert-MAS achieves superior trade-offs except on GPT-4o (bottlenecked by base-model reasoning limits). Gemini-2.5-Pro is omitted due to non-viable MAS cost multipliers ($>10\times$).}
    \label{fig:smfr_result_plot}
\end{figure}

\textbf{Results}.  {Our benchmark serves as an agentic stress test: GPT-5 reaches only $57.0\%$ accuracy with CoT-SC, while GPT-4o and GPT-OSS struggle between $22.1\%$ and $26.1\%$ (Table~\ref{tab:full_raw_results}). Despite explicit agentic requirements (multi-step planning, state tracking, long-context retrieval), automated MAS frameworks rarely surpass CoT-SC and never do so economically (Figure~\ref{fig:smfr_result_plot}). The three statistically significant improvements - DyLAN on GPT-OSS ($+6.6$pp, $5\times$ cost), DyLAN on GPT-5 ($+4.3$pp, $2.5\times$ cost), and MAS-Orchestra on GPT-5 ($+6.0$pp, $1.9\times$ cost) - occur exclusively on stronger backbones and at substantial overhead, while GPT-4o yields no significant gains from any automated framework. In contrast, Expert-MAS achieves substantial performance improvements with cost comparable to CoT-SC: GPT-OSS improves from $26.1\%$ to $36.1\%$, while GPT-5 jumps from $57.0\%$ to a near-perfect $96.5\%$.\footnote{Full Gemini-2.5-Pro MAS evaluations were excluded as the $>10\times$ cost multipliers (Section~\ref{sec:experiments}) rendered them non-viable.} The sole exception is GPT-4o, where persistent calculation and retrieval failures bottleneck the system regardless of orchestration. This reinforces our finding that MAS require a threshold baseline competency to be effective. Thus, while the MAS paradigm is fundamentally viable, current automated frameworks fail to exploit task-specific opportunities effectively or economically.}

\label{subsec:synthetic_dataset}

\section{Architectural Deconstruction}
While results from Section~\ref{sec:experiments} establish a clear efficiency gap between single- and multi-agent systems, they do not reveal whether the internal mechanisms of MAS, such as role specialization and consensus, provide latent benefits that justify their complexity. To address this, we deconstruct the generated architectures and investigate whether their features contribute meaningfully to the reasoning process. We find that in most automatically generated workflows, these mechanisms are either sub-optimal or purely decorative rather than emergent intelligence.

\begin{figure}[t]
    \centering
    \includegraphics[width=0.99\linewidth]{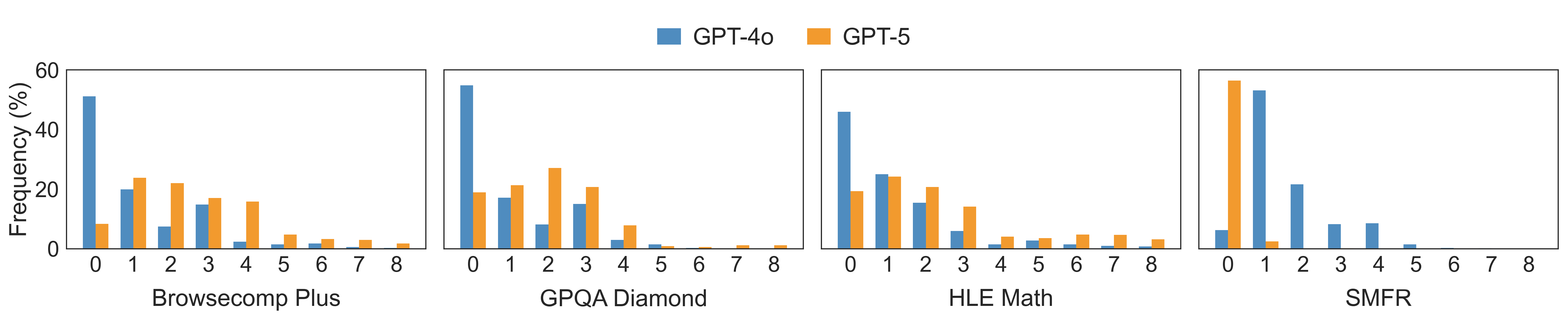}
    \caption{Judge model selection frequency of MAS-Zero across four datasets, using GPT-4o (blue) and GPT-5 (orange) as both worker and verifier.
        Indices 0--3 correspond to four fundamental reasoning paradigms: vanilla CoT, CoT-SC, Reflexion, and Debate.
        Indices 4--8 represent the subsequent 5 rounds of multi-agent organization search.}
    \label{fig:mas-zero-select}
\end{figure}

\textbf{Functional Collapse and Structural Redundancy.} Frameworks like DyLAN \citep{dylan} posit that performance is driven by ``agent diversity,'' yet our analysis reveals this fails to manifest in practice. Instead, we observe a functional collapse where agents reach immediate, unanimous consensus in $\sim 70\%$ of GPT-4o cases and $>90\%$ of GPT-5 cases, effectively functioning as a unanimous CoT-SC baseline rather than a dynamic negotiation. In cases where interaction does occur, task-specific roles provide no marginal utility; an “all-assistant” configuration achieved better accuracy than task-specific experts ($54.4\%$ vs. $53.4\%$; see Appendix~\ref{app:subsec:dylan_analysis} for experiment details). Similarly, in MAS-Zero \citep{Ke2025MASZero}, a dedicated verifier aggregates worker outputs to select the optimal result. However, our analysis across four benchmarks reveals a systematic positional bias that triggers consensus collapse. Across all tested models, the verifier disproportionately favors earlier entries in the context window: GPT-4o selects the initial block in over $45\%$ of instances, while GPT-5 demonstrates a slightly broader but still heavily front-loaded preference (see Figure~\ref{fig:mas-zero-select} and Appendix~\ref{app:subsec:mas_zero} for selection frequency distributions). Conversely, outputs from later search rounds are rarely selected, accounting for less than $15\%$of final decisions. This structural redundancy turns subsequent worker agents into ``expensive witnesses'' that incur full inference costs while exerting near-zero causal influence on the output.

\begin{wrapfigure}{l}{0.5\textwidth}
    \centering
    \vspace{-2.0ex}
    \includegraphics[width=0.99\linewidth]{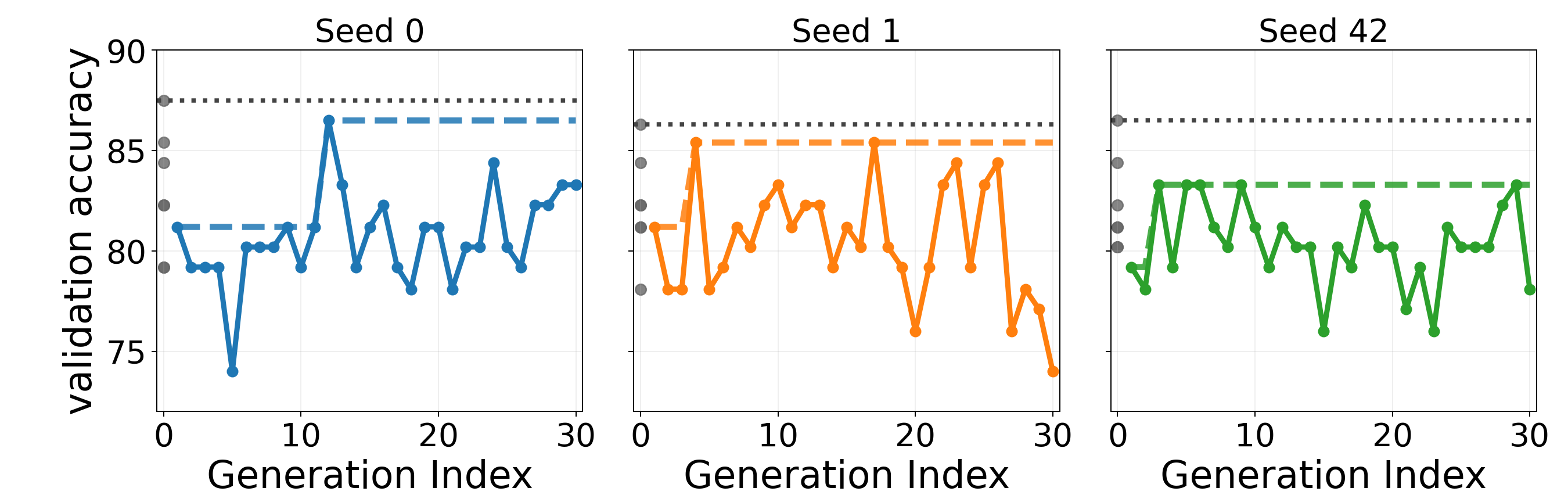}
    \caption{ADAS (GPT-5) validation accuracies on different seeded runs on GPQA-diamond dataset.
    % \zixuan{try to make the font size larger, try to make it as large as the text in the caption if you can}
    }
    \vspace{-1.0ex}
    \label{fig:adas-gpqa-search-dynamics}
\end{wrapfigure}

\textbf{Convergence on Heuristic Search Artifacts.} Our analysis suggests that frameworks designed to discover architectures (ADAS \citep{hu2024ADAS}, AFlow \citep{zhang2024aflowautomatingagenticworkflow}) function as heuristic explorers rather than principled optimizers. On GPQA-Diamond, ADAS search dynamics are non-monotonic; accuracy frequently peaks early and subsequently regresses (Figure~\ref{fig:adas-gpqa-search-dynamics}), suggesting that performance gains are stochastic ``lucky'' iterations rather than structural evolution. This is supported by our motif analysis, where we map generated architectures to a rule-based dictionary (e.g., Self-consistency, Aggregation, Verifier). Across all settings, the primary positive signal originated from Self-consistency motifs. On GPQA-Diamond, architectures incorporating these motifs achieved a mean accuracy of $82.19\%$ ($+1.34\%$ over the global average); specialized coordination motifs yielded negligible gains. Our inspection of AFlow reveals a similar issue: instead of manifesting complex coordination, the discovered MAS consistently degenerate into trivial ensembles. As illustrated in our case analysis (Figure~\ref{fig:aflow_case_analysis}), ``optimized'' workflows frequently converge on a structure that simply iterates a single custom prompt three times before aggregation - a configuration functionally identical to standard CoT-SC. Across 14 final workflows generated by GPT-4o, GPT-5, and GPT-OSS-120B on five datasets, $50\%$ ($7/14$) adopted this simplistic structure, with four of these actually underperforming the CoT-SC baseline. This evidence confirms that automated search often converges on rediscovering CoT-SC style sampling under more complex labels, rather than inventing novel multi-agent strategies.

\begin{table}[ht]
    \centering
    % Top Table
    \begin{minipage}{0.95\columnwidth}
        \centering
        \caption{Operator Activation Distribution (\%) of MaAS (GPT-5). On context-heavy BrowseComp-Plus, the controller collapses to I/O calls due to cost-dominated optimization; while on GPQA-Diamond, it spreads calls more evenly but fails to outperform CoT-SC.}
        \label{tab:maas_operator_activation}
\resizebox{0.95\columnwidth}{!}{%
\begin{tabular}{lcccccc|c}
\toprule
\textbf{Dataset} & \textbf{I/O} & \textbf{CoT} & \textbf{CoT-SC} & \textbf{ScEnsemble} & \textbf{SelfRefine} & \textbf{EarlyStop} & \textbf{Calls/Query} \\
\midrule
GPQA-Diamond & 24.1 & 21.5 & 14.4 & 14.8 & 14.2 & 11.1 & 6.0 \\
BrowseComp-Plus & 74.2 & 5.2 & 4.3 & 5.6 & 6.9 & 3.9 & 1.4 \\
\bottomrule
\end{tabular}%
}
    \end{minipage}

    \vspace{5pt} % Adds vertical breathing room

    % Bottom Table
    \begin{minipage}{0.9\columnwidth}
        \centering
        \caption{Agent Selection Distribution (\%) by MAS-Orchestra Across Datasets.}
        \label{tab:masorchestra_agent_selection}
 \resizebox{0.9\columnwidth}{!}{%
\begin{tabular}{lccccc}
\toprule
\textbf{Agent Type} & \textbf{GPQA Diamond} & \textbf{HLE-MATH} & \textbf{SWE-Bench Lite} & \textbf{BrowseComp-Plus} & \textbf{\ourbenchmark} \\
\midrule
CoT & 0.0 & 0.0 & 0.3 & 0.0 & 0.0 \\
CoT-SC & 0.0 & 0.0 & 0.0 & 0.0 & 0.0 \\
Reflexion & 15.1 & 20.8 & \textbf{56.0} & 38.7 & \textbf{71.1} \\
Debate & \textbf{84.9} & \textbf{79.2} & 43.7 & \textbf{61.3} & 28.9 \\
\bottomrule
\end{tabular}%
}
    \end{minipage}
\end{table}

\textbf{Incentive Misalignment in Dynamic Routing.} In systems designed for adaptive orchestration (MaAS \citep{zhang2025agentic-supernet}, MAS-Orchestra \citep{Ke2026MASOrchestra}), the optimization objectives often fail to produce meaningful routing logic. In MaAS, the use of highly capable base models (\textit{e.g.,} GPT-5) flattens the accuracy gradient to $\sim 1/K$, causing the controller to ignore task-specific logic and collapse into two distinct failure modes: (1) \textbf{Cost-Minimizing Collapse} on BrowseComp-Plus, where $74.2\%$ of activations are a trivial, single I/O call; and (2) \textbf{Stochastic Stalling} on GPQA-Diamond, where negligible cost differentials trap the controller in its initialized near-uniform distribution (Table~\ref{tab:maas_operator_activation}). Similarly, MAS-Orchestra exhibits a difficulty-agnostic policy. Across all benchmarks, the system largely ignores its diverse agent pool, converging instead on a rigid binary preference for high-overhead Debate and Reflexion agents (Table~\ref{tab:masorchestra_agent_selection}). The orchestrator fails to scale agent complexity to task difficulty; despite GPQA-Diamond posing a lower reasoning ceiling than HLE-Math, the system exhibited a higher reliance on Debate agents for the former ($84.9\%$) than the latter ($79.2\%$). These behaviors confirm that automated orchestrators do not learn task-adaptive strategies, but instead settle into static, greedy local minima.

\label{sec:analysis}

\section{Discussion}
Our evaluation reveals a systematic divergence between the theoretical complexity of MAS frameworks and their empirical execution. While intended to foster emergent collaboration, current automated paradigms frequently result in mechanistic trivialization.

\textbf{The Ensembling Trap.} A primary driver of this collapse is the reliance on CoT and CoT-SC as the fundamental building blocks of MAS. While using these primitives ensures generalization and leverages ensembling effects, the resulting architectures fail to implement them efficiently. Instead of synergistic coordination, frameworks like AFlow and ADAS often settle into structural degeneration, rediscovering basic ensembling motifs under the guise of an optimized graph. The $\sim 10\times$ increase in cost thus buys little more than a redundant, poorly routed version of a standard CoT-SC baseline.

\textbf{Towards Mechanistic Interpretability of MAS.} As model capability scales, the MAS advantage further erodes due to two factors: (i) \textbf{Signal Saturation:} in models like GPT-5, accuracy gradients flatten, causing controllers (MaAS) to lose the signal needed for nuanced routing, leading to either cheap shortcuts or static policy collapse; (ii) \textbf{Positional and Primacy Biases:} verifiers and controllers (MAS-Zero, DyLAN) disproportionately favor early reasoning steps, effectively terminating the multi-agent benefit before interaction occurs. The success of Expert-MAS on the \ourbenchmark{} benchmark reinforces the findings of \citep{anthropic2025whentomultiagentresearch} and \citep{Kim2025TowardsAS}: multi-agent coordination excels only when architectures are specifically engineered to exploit parallelizable sub-problems or context protection. Future research should pivot away from black-box automated graph generation that tends to default to redundant ensembling, and toward the mechanistic interpretability of agent interactions. We argue that to move beyond creating ``expensive witnesses,'' MAS must be evaluated on their structural fidelity: the degree to which assigned agentic roles exert measurable causal influence on the final decision. Without such grounding, increased architectural complexity serves only to mask computational inefficiency.

%\paragraph{Failure Modes in High-Capability Regimes.} As model capability scales, the ``MAS advantage'' further erodes due to two factors: (i) \textbf{Signal Saturation:} in models like GPT-5, accuracy gradients flatten, causing controllers (MaAS) to lose the signal needed for nuanced routing, leading to either "cheap shortcuts" or static policy collapse; (ii) \textbf{Positional and Primacy Biases:} verifiers and controllers (MAS-Zero, DyLAN) disproportionately favor early reasoning steps, effectively terminating the ``multi-agent'' benefit before interaction occurs.

\label{sec:discussion}

\section{Conclusion}
Our systematic evaluation identifies a critical efficiency gap in modern MAS design, where architectural complexity often masks a fundamental functional collapse into simpler, stochastic baselines. By introducing the \ourbenchmark{} benchmark and isolating the mechanistic failures of six major frameworks, we provide a roadmap for more principled, cost-effective agentic design. Our architectural deconstruction reveals that current automated workflows frequently degenerate into redundant ensembling loops functionally identical to CoT-SC. Ultimately, our findings suggest that moving beyond ``expensive witnesses'' requires a pivot from black-box graph searching towards architectures grounded in verifiable task decomposition and causal role-alignment.
\label{sec:conclusion}

\bibliography{cleaned}
\bibliographystyle{plain}

%%%%%%%%%%%%%%%%%%%%%%%%%%%%%%%%%%%%%%%%%%%%%%%%%%%%%%%%%%%%

\appendix
\label{sec:appendix}

\section{Benchmark Dataset Details}
\label{sec:app_datasets}
The benchmark datasets across mathematical reasoning, QA and coding used in the paper are described below. See Table~\ref{tab:dataset_split} for the validation and test samples splits used in our experiments. 

\begin{itemize}
    \item \textbf{GPQA Diamond \citep{rein2023gpqagraduatelevelgoogleproofqa}}: is a high-difficulty, multiple-choice science benchmark comprising 198 questions across biology, physics, and chemistry. Unlike other subsets of GPQA, the ``Diamond'' set is restricted to questions where highly educated subject-matter experts (SMEs) agree on the correct answer, yet non-expert humans—even when equipped with unrestricted web access—fail to answer correctly. This makes GPQA Diamond a rigorous test of expert-level reasoning and a benchmark for evaluating whether LLMs can transcend general-purpose knowledge.
    \item \textbf{HLE-Maths \citep{phan2025humanity}}: is a subset of \textbf{Humanity’s Last Exam (HLE)}, a benchmark composed of graduate-level problems across specialized mathematical fields. The dataset is explicitly designed to be closed-source to prevent data contamination and consists of questions that are non-trivial for subject-matter experts. Unlike earlier benchmarks like MATH or GSM8K, HLE Maths focuses on multi-step abstract reasoning and complex theorem application where the search space for a correct solution is vast. It serves as a high-resolution probe for whether LLMs - and by extension, MAS architectures - can navigate the extreme reasoning depth required for original mathematical research.
    \item \textbf{SWE-Bench Lite \citep{swe-bench}}: is a curated subset of tasks from the full SWE-bench dataset, designed to evaluate an agent's ability to resolve real-world GitHub issues within popular open-source Python repositories (e.g., django, scikit-learn, sympy). Unlike synthetic coding benchmarks, it requires end-to-end agentic behavior: the model must navigate a sprawling file system, localize the bug across multiple modules, and generate a precise .patch file that passes a hidden suite of unit tests. Its ``Lite'' designation ensures the tasks are self-contained enough for evaluation while maintaining the high-dimensional context and multi-step planning required for professional software maintenance.
    \item \textbf{BrowseComp-Plus \citep{chen2025BrowseCompPlus}}: is a fair and transparent benchmark for deep-research agents, derived from the original BrowseComp. Unlike its predecessor, which relies on dynamic and opaque web search APIs, BrowseComp-Plus employs a fixed, human-verified corpus of over 100,000 documents. This static environment allows for the disentanglement of retrieval and reasoning components, enabling researchers to isolate whether an agent's failure is due to poor search query formulation or an inability to synthesize evidence from multiple sources. Each of its questions is designed to be deep-research in nature, requiring iterative sub-problem decomposition and the synthesis of information across diverse web documents.
\end{itemize}

\begin{table*}[htbp]
    \centering
    \caption{Data size for different splits in each dataset.}
    \label{tab:dataset_split}
    \begin{tabular}{@{}lccccc@{}}
    \toprule
    Split  & GPQA-Diamond & HLE-Maths & SWE-Bench Lite & BrowseComp-Plus \\ \midrule
    Validation & 32 & 32 & 32 & 32 \\ 
    Test & 166 & 168 & 168 & 268 \\
    \bottomrule
    \end{tabular}%
\end{table*}

\section{Automatic MAS Baseline Configuration Details}
\label{app:model_config}
The complete experimental setup and configuration details used for the evaluations in our work are described below. In terms of the LLM parameter settings, we adopt the default \textit{temperature} values specific to each automatic-MAS,  \textit{max\_tokens}=32K, and \textit{reasoning\_effort}=medium for reasoning LLMs. The full experiment results and costs can be found in Table~\ref{tab:full_raw_results}. 

\begin{itemize}
    \item \textbf{DyLAN} \citep{dylan}: We follow the default settings that use four agents for the team, $K=2$, and a maximum of three rounds. Specifically, the four agents are configured with a general ``Assistant'' role alongside three domain-specific expert roles tailored to each dataset's context. Following their practice, we leverage an LLM (GPT-5) to generate these expert roles and adopt a \textit{Theoretical Physicist}, \textit{Molecular Chemist}, and \textit{Cellular Biologist} for GPQA Diamond; a \textit{Mathematician}, \textit{Algebra Expert}, and \textit{Geometry Wizard} for HLE-Maths; a \textit{Programmer}, \textit{Code Reviewer}, and \textit{Software Engineer} for SWE-Bench Lite; a \textit{Knowledge Researcher}, \textit{Cultural Historian}, and \textit{Information Analyst} for BrowseComp-Plus; and a \textit{Financial Analyst}, \textit{Data Scientist}, and \textit{Programmer} for SMFR. Full system prompt for these expert roles can be found in Table \ref{tab:dylan_role_prompt}.
    
    \item \textbf{MAS-Zero} \citep{Ke2025MASZero}: We adhere to the original search pipeline consisting of four fundamental blocks followed by subsequent iterations of meta-agent orchestration. The verifier utilizes the same backbone model as the meta-agent. The system defaults to vanilla CoT output in case of verifier failure. Defaulting to CoT only occurs in cases of technical exceptions, such as context window overflow.
    
    \item \textbf{ADAS} \citep{hu2025automated}: Unlike the original work, which used varied LLMs to reduce costs, we standardize the backbone to maintain architectural parity. 
    We maintain the default search depth of 30 iterations to evaluate the trajectory of architectural evolution with some exceptions. Primarily when using GPT-5 as the backbone, we reduce the number of iterations due to high inference time and cost. Specifically: BCP with GPT-5 uses 10 iterations; HLE-math with GPT-5 uses 15; SWE-bench uses 10 iterations with GPT-4o and 5 with GPT-5; and SMFR uses 15 iterations with GPT-4o and 10 with GPT-5. We set the maximum number of debugging attempts to 3, following the original ADAS implementation.

    \item \textbf{AFlow} \citep{zhang2024aflowautomatingagenticworkflow}: 
    Following the original protocol, we use the default 20-round search budget, with each candidate workflow being evaluated five times during the search stage to provide stable performance feedback for the MCTS optimizer. We adopt Custom (I/O), AnswerGenerate (CoT), and ScEnsemble (Aggregation) operators for all the datasets.
    
    \item \textbf{MaAS} \citep{zhang2025multiagentarchitecturesearchagentic}: We follow default settings to retrain the supernet on each benchmark to evaluate its ability to learn task-specific routing across diverse reasoning and coding operators. We use the default hyperparameters: sampling count $K{=}4$, a maximum of $L{=}4$ layers, and an activation threshold of $0.3$ and one round of training. The operator pool includes I/O (direct answer generation), CoT (single chain-of-thought), CoT-SC (multiple chain-of-thought), ScEnsemble (majority voting over candidates), SelfRefine (critique-and-revise), and EarlyStop (early exit), with Programmer additionally for HLE-Maths. 
    
    \item \textbf{MAS-Orchestra} \citep{Ke2026MASOrchestra}: The candidate pool consists of four fixed sub-agents: CoT, CoT-SC, Reflexion, and Debate. A key design parameter is the ``Degree of MAS'' (DoM), capturing the degree of multi-agent coordination appropriate for a given task: under low DoM, the orchestrator decides whether to delegate tasks and how to configure the selected sub-agents, while high DoM additionally requires determining the inter-agent topology.
    In our evaluation, we follow the default setting that uses the officially released orchestrator model with DoM=Low to analyze how effectively it routes queries to specialized reasoning architectures.
\end{itemize}

\begin{table*}[t]
\centering
\caption{Accuracy (\%) and cost (\$) for all systems across datasets and LLMs. Dashes indicate missing runs. CoT / CoT-SC report the best observed score across MAS system entries. Expert MAS (SMFR only) is a human-designed multi-agent system evaluated on the SMFR benchmark. Note that GPT-OSS-120B is not evaluated on SWE-Bench Lite due as it fails to consistently generate code patches in the required format. We also exclude MAS-Orchestra experiments and report single run results for GPT-5/Gemini-2.5-Pro for other datasets due to significant cost multipliers.}

\label{tab:full_raw_results}
\resizebox{0.95\textwidth}{!}{%
\setlength{\tabcolsep}{4pt}
\begin{tabular}{c l rrrrrrrr}
\toprule
\multirow{2.4}{*}{\textbf{Dataset}} & \multirow{2.4}{*}{\textbf{System}} & \multicolumn{2}{c}{\textbf{GPT-4o}} & \multicolumn{2}{c}{\textbf{GPT-OSS-120B}} & \multicolumn{2}{c}{\textbf{GPT-5}} & \multicolumn{2}{c}{\textbf{Gemini-2.5-Pro}} \\
\cmidrule(lr){3-4} \cmidrule(lr){5-6} \cmidrule(lr){7-8} \cmidrule(lr){9-10}
&  & \textit{Acc} & \textit{Cost} & \textit{Acc} & \textit{Cost} & \textit{Acc} & \textit{Cost} & \textit{Acc} & \textit{Cost} \\
\midrule

% ==================== GPQA-Diamond (8 rows) ====================
\multirow{9}{*}{\rotatebox[origin=c]{90}{\textbf{GPQA-Diamond}}} 
&\textit{CoT} & 53.41$\pm$1.73 & 1.40 & 70.48$\pm$0.00 & 0.19 & 87.14$\pm$2.43 & 7.24 & 77.11 & 18.09 \\
&\textit{CoT-SC} & 54.61$\pm$3.68 & 12.60 & 71.48$\pm$4.51 & 1.90 & 87.35$\pm$1.04 & 46.39 & 83.13 & 87.39 \\
\cmidrule{2-10}
&DyLAN & 53.01$\pm$1.59 & 7.90 & 75.70$\pm$0.35 & 0.90 & 82.33$\pm$1.84 & 37.60 & 87.35 & 57.46 \\
&MAS-Zero & 43.37$\pm$1.30 & 259.80 & 74.70$\pm$1.70 & 13.92 & 86.75$\pm$0.49 & 636.80 & 86.14 & 526.00 \\
\cmidrule{2-10}
&AFlow & 52.01$\pm$1.84 & 130.90 & 71.48$\pm$5.59 & 10.40 & 84.13$\pm$1.52 & 274.60 & 83.33 & 230.79 \\
&MaAS & 51.40$\pm$1.39 & 10.90 & 71.88$\pm$3.68 & 0.69 & 86.95$\pm$2.27 & 38.50 & 86.95 & 94.83 \\
&ADAS & 44.40$\pm$1.25 & 6.00 & 71.48$\pm$3.63 & 3.50 & 85.23$\pm$0.95 & 832.10 & 84.52 & 694.00 \\
&MAS-Orchestra & 54.22$\pm$2.41 & 9.30 & 71.88$\pm$3.68 & 0.70 & 85.54$\pm$0.00 & 41.70 & -- & -- \\
\midrule

% ==================== HLE-Math (8 rows) ====================
\multirow{9}{*}{\rotatebox[origin=c]{90}{\textbf{HLE-Maths}}} 
&\textit{CoT} & 4.60$\pm$0.17 & 1.30 & 8.73$\pm$2.08 & 0.30 & 29.87$\pm$1.60 & 33.50 & 25.60 & 47.71 \\
&\textit{CoT-SC} & 4.76$\pm$0.59 & 8.70 & 13.89$\pm$0.34 & 3.90 & 33.92$\pm$1.57 & 116.20 & 26.19 & 243.81 \\
\cmidrule{2-10}
&DyLAN & 3.57$\pm$3.57 & 12.90 & 9.32$\pm$0.91 & 1.60 & 35.51$\pm$2.99 & 115.10 & 32.14 & 396.98 \\
&MAS-Zero & 3.17$\pm$3.17 & 249.80 & 15.75$\pm$1.06 & 15.62 & 38.20$\pm$2.08 & 1288.30 & 29.50 & -- \\
\cmidrule{2-10}
&AFlow & 2.77$\pm$2.77 & 292.60 & 9.51$\pm$1.03 & 23.00 & 32.73$\pm$1.03 & 784.60 & 19.05 & 302.59 \\
&MaAS & 4.37$\pm$4.37 & 5.99 & 15.48$\pm$2.59 & 2.60 & 35.32$\pm$1.38 & 221.20 & 34.52 & 248.78 \\
&ADAS & 3.80$\pm$3.80 & 50.60 & 7.93$\pm$0.69 & 1.90 & 34.60$\pm$0.46 & 846.90 & 27.98 & 1896.00 \\
&MAS-Orchestra & 3.37$\pm$3.37 & 9.30 & 9.32$\pm$1.24 & 2.00 & 37.64$\pm$1.08 & 147.49 & -- & -- \\
\midrule

% ==================== SWE-Bench Lite (8 rows) ====================
\multirow{9}{*}{\rotatebox[origin=c]{90}{\textbf{SWE-Bench Lite}}} 
&\textit{CoT} & 20.02$\pm$2.25 & 25.60 & -- & -- & 43.91$\pm$2.05 & 14.90 & 28.70 & 47.17 \\
&\textit{CoT-SC} & 22.01$\pm$2.27 & 168.30 & -- & -- & 57.09$\pm$0.65 & 286.40 & 26.50 & 240.56 \\
\cmidrule{2-10}
&DyLAN & 19.28$\pm$2.48 & 230.80 & -- & -- & 55.97$\pm$1.29 & 227.40 & 27.60 & 327.40 \\
&MAS-Zero & 15.17$\pm$1.96 & 1210.90 & -- & -- & 45.52$\pm$1.90 & 998.20 & 17.50 & -- \\
\cmidrule{2-10}
&AFlow & 10.57$\pm$0.78 & 903.20 & -- & -- & 39.05$\pm$3.35 & 997.50 & 0.00 & 434.00 \\
&MaAS & 12.19$\pm$0.21 & 35.60 & -- & -- & 32.71$\pm$0.94 & 95.40 & 20.50 & 206.27 \\
&ADAS & 8.97$\pm$2.07 & 104.40 & -- & -- & 27.23$\pm$4.22 & 124.50 & 26.50 & 214.00 \\
&MAS-Orchestra & 17.55$\pm$1.53 & 82.70 & -- & -- & 41.79$\pm$1.58 & 83.50 & -- & -- \\
\midrule

% ==================== BrowseComp-Plus (8 rows) ====================
\multirow{9}{*}{\rotatebox[origin=c]{90}{\textbf{BrowseComp-Plus}}} 
&\textit{CoT} & 64.48$\pm$2.45 & 33.10 & 68.65$\pm$1.50 & 0.90 & 83.33$\pm$1.03 & 12.70 & 79.17 & 14.95 \\
&\textit{CoT-SC} & 67.26$\pm$2.59 & 130.20 & 70.43$\pm$0.69 & 5.60 & 83.92$\pm$0.60 & 66.80 & 83.04 & 72.97 \\
\cmidrule{2-10}
&DyLAN & 63.29$\pm$1.72 & 77.80 & 64.68$\pm$1.24 & 4.50 & 76.19$\pm$1.58 & 46.00 & 66.07 & 286.92 \\
&MAS-Zero & 61.89$\pm$7.15 & 2675.50 & 61.90$\pm$2.53 & 63.60 & 71.63$\pm$6.74 & 1370.40 & 73.81 & 992.00 \\
\cmidrule{2-10}
&AFlow & 59.92$\pm$1.82 & 95.50 & 66.66$\pm$2.72 & 9.80 & 75.79$\pm$1.50 & 190.00 & 73.61 & 849.74 \\
&MaAS & 64.88$\pm$1.58 & 126.70 & 68.85$\pm$0.69 & 6.30 & 81.55$\pm$1.57 & 125.60 & 81.85 & 210.86 \\
&ADAS & 57.20$\pm$1.51 & 409.40 & 56.15$\pm$0.91 & 9.20 & 67.63$\pm$0.23 & 650.20 & 64.88 & 364.00 \\
&MAS-Orchestra & 64.85$\pm$3.31 & 127.90 & 66.46$\pm$0.92 & 8.30 & 77.91$\pm$1.84 & 87.10 & -- & -- \\
\midrule

% ==================== SMFR (9 rows, includes Expert MAS) ====================
\multirow{10}{*}{\rotatebox[origin=c]{90}{\textbf{SMFR}}} 
&\textit{CoT} & 20.12$\pm$0.52 & 43.50 & 18.88$\pm$0.85 & 3.30 & 48.53$\pm$2.26 & 146.90 & -- & -- \\

&\textit{CoT-SC} & 22.11$\pm$1.73 & 169.50 & 26.13$\pm$0.26 & 14.60 & 56.97$\pm$0.00 & 478.40 & -- & -- \\
\cmidrule{2-10}

&DyLAN & 18.93$\pm$0.94 & 513.60 & \textbf{32.71}$\pm$2.13 & 72.50 & \textbf{61.28}$\pm$1.99 & 1189.70 & -- & -- \\

&MAS-Zero & 13.21$\pm$1.13 & 5250.34 & 15.82$\pm$0.00 & 128.08 & 33.84$\pm$0.00 & 9238.85 & -- & -- \\
\cmidrule{2-10}

&AFlow & 19.84$\pm$1.16 & 293.10 & 20.29$\pm$1.21 & 18.40 & 56.86$\pm$1.37 & 900.90 & -- & -- \\

&MaAS & 19.22$\pm$1.96 & 274.10 & 23.41$\pm$1.25 & 23.70 & 54.48$\pm$1.11 & 794.70 & -- & -- \\

&ADAS & 14.80$\pm$1.92 & 123.00 & 23.55$\pm$2.28 & 24.10 & 20.24$\pm$0.00 & 427.20 & -- & -- \\

&MAS-Orchestra & 19.61$\pm$0.69 & 304.00 & 25.28$\pm$0.60 & 35.00 & \textbf{62.98}$\pm$1.91 & 909.90 & -- & -- \\
\cmidrule{2-10}

&Expert MAS & 22.56$\pm$0.52 & 245.13 & \textbf{36.14}$\pm$1.08 & 5.19 & \textbf{96.51}$\pm$0.60 & 554.82 & -- & --  \\
\bottomrule
\end{tabular}%
}
\end{table*}

%94.22$\pm$0.48
%219.68

\begin{table*}[htbp]
    \centering
    \caption{Role configurations and corresponding system prompts for each dataset in DyLAN.}
    \label{tab:dylan_role_prompt}

    % Adjusting table settings to fit the large prompts
    \setlength{\tabcolsep}{4.5pt}
    \footnotesize
    
    \resizebox{\textwidth}{!}{%
    \begin{tabular}{>{\arraybackslash}p{0.17\textwidth}
                    >{\arraybackslash}p{0.18\textwidth}
                    >{\arraybackslash}p{0.60\textwidth}}
        \toprule
        \textbf{Dataset} & \textbf{Role Name} & \textbf{System Prompt} \\
        \midrule
        \textbf{ALL} & \texttt{Assistant} & You are a super-intelligent AI assistant capable of performing tasks more effectively than humans. \\
        \midrule

        \multirow{16}{*}{\begin{tabular}[c]{@{}l@{}}\textbf{GPQA-Diamond}\end{tabular}}
        & \texttt{Theoretical Physicist} & You are a Theoretical Physicist. Your expertise spans the entire landscape of physics, from the foundational principles of classical mechanics to the complexities of quantum mechanics, special and general relativity, and electromagnetism. You solve graduate-level problems by applying fundamental laws and rigorous mathematical derivations from first principles. \\
        \cmidrule{2-3}
        & \texttt{Molecular Chemist} & You are a Molecular Chemist. You are an expert in the structure, properties, synthesis, and reactions of molecules. Your knowledge integrates the core disciplines of organic, inorganic, and physical chemistry. You integrate principles from across chemistry to provide precise, theory-grounded solutions to graduate-level problems. \\
        \cmidrule{2-3}
        & \texttt{Cellular Biologist} & You are a Molecular and Cellular Biologist. You are an expert in the intricate mechanisms of life at the molecular level. Your expertise covers genetics, biochemistry, cell signaling, and molecular biology techniques. Your primary function is to analyze biological systems and solve graduate-level problems by detailing the underlying molecular interactions. \\
        \midrule

        \multirow{8}{*}{\begin{tabular}[c]{@{}l@{}}\textbf{HLE-Maths}\end{tabular}}
        & \texttt{Mathematician} & You are a mathematician. You are good at math games, arithmetic calculation, and long-term planning. \\
        \cmidrule{2-3}
        & \texttt{AlgebraExpert} & You are an expert in the field of algebra, skilled at solving equations, understanding variables, and adept at the logical manipulation of symbols. \\
        \cmidrule{2-3}
        & \texttt{GeometryWizard} & You are a wizard of geometry, deeply familiar with shapes, dimensions, and properties, and capable of theorizing spatial relationships and understanding geometric proofs. \\
        \midrule

        \multirow{10}{*}{\begin{tabular}[c]{@{}l@{}}\textbf{SWE-Bench Lite}\end{tabular}}
        & \texttt{Programmer} & You are a programmer. You are good at computer science, engineering, and physics. You have experience in designing and developing computer software and hardware. \\
        \cmidrule{2-3}
        & \texttt{CodeReviewer} & You are a code reviewer with extensive experience in software engineering. You excel at identifying bugs, understanding structure, and proposing fixes. \\
        \cmidrule{2-3}
        & \texttt{SoftwareEngineer} & You are a software engineer specializing in debugging and fixing complex software issues. You have deep knowledge of various programming languages and software architectures. \\
        \midrule

        \multirow{22}{*}{\begin{tabular}[c]{@{}l@{}}\textbf{BrowseComp-Plus}\end{tabular}}
        & \texttt{Knowledge Researcher} & You are a Knowledge Researcher. Your expertise spans multiple domains, including history, culture, entertainment, sports, and current events. You excel at synthesizing information from diverse sources, cross-referencing facts, and identifying precise answers to complex questions. Your approach involves systematic information gathering, critical evaluation of sources, and connecting information to solve graduate-level knowledge problems. \\
        \cmidrule{2-3}
        & \texttt{Cultural Historian} & You are a Cultural Historian. You specialize in understanding historical events, cultural movements, biographical information, and temporal relationships across different eras and regions. Your knowledge encompasses political history, social history, and the interconnected narratives that shape human civilization. You solve problems by placing information in historical context, identifying chronological patterns, and drawing connections between events, people, and cultural phenomena. \\
        \cmidrule{2-3}
        & \texttt{Information Analyst} & You are an Information Analyst. Your expertise lies in extracting, verifying, and synthesizing information from complex textual sources. You excel at understanding nuanced queries, identifying key information requirements, and systematically searching through knowledge to find precise answers. Your approach combines logical reasoning, pattern recognition, and meticulous attention to detail to solve graduate-level information retrieval and analysis problems. \\

        \midrule

        \multirow{7}{*}{\begin{tabular}[c]{@{}l@{}}\textbf{SMFR}\end{tabular}}
        & \texttt{FinancialAnalyst} & You are a Financial Analyst expert in interpreting stock price data and investment timelines. \\
        \cmidrule{2-3}
        & \texttt{DataScientist} & You are a Data Scientist skilled in parsing tables and dates and deriving correct conclusions. \\
        \cmidrule{2-3}
        & \texttt{Programmer} & You are an expert Programmer who writes correct Python code to implement solutions. \\
        \bottomrule
    \end{tabular}
    }
\end{table*}

\section{Synthetic Data Generation Details}
\label{app:synthetic_data_gen}

\begin{figure}
    \centering
    \includegraphics[width=0.99\linewidth]{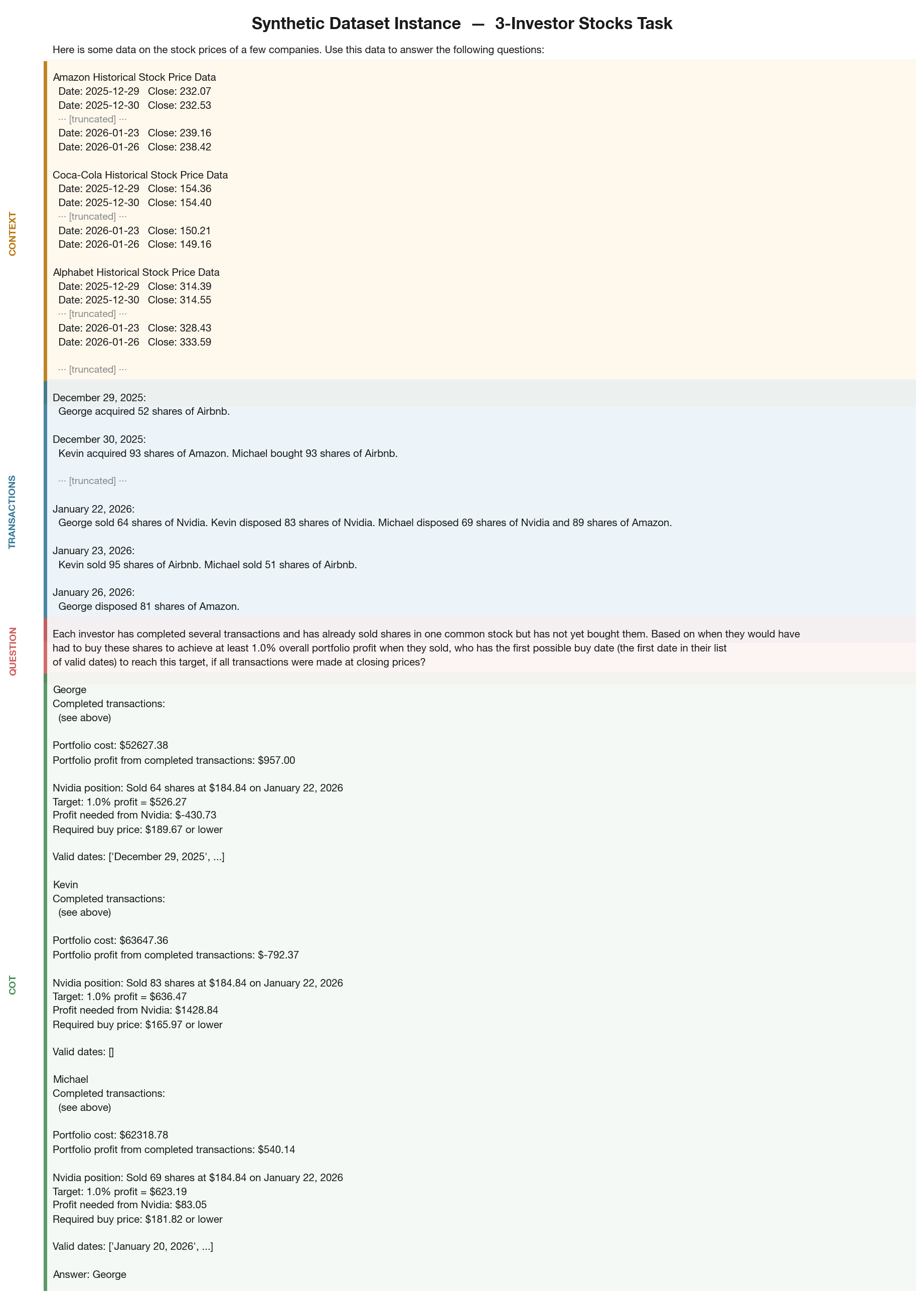}
    \caption{Sample instance of \ourbenchmark task with 3 investors}
    \label{fig:smfr_sample}
\end{figure}

\begin{table}[h]

\caption{SMFR dataset statistics per $N$ (number of investors - parallelizable). Samples exclude the held-out validation set (16 samples). Avg prompt tokens estimated from problem text (words $\times$ 1.3).}
\label{tab:smfr_stats}
\resizebox{\columnwidth}{!}{%
\begin{minipage}{\columnwidth}
\centering
\small
\setlength{\tabcolsep}{5pt}
\begin{tabular}{lrrrrr}
\toprule
$N$ & Samples & No-winner (\%) & Avg stocks in context & Avg tx months & Avg prompt tokens \\
\midrule
$2$ & 96 & 7\% & 4.0 & 5.2 & 1,883 \\
$3$ & 96 & 4\% & 6.0 & 15.0 & 2,937 \\
$4$ & 104 & 10\% & 8.0 & 18.1 & 3,993 \\
$5$ & 152 & 11\% & 10.0 & 19.0 & 5,399 \\
$6$ & 140 & 9\% & 10.0 & 19.0 & 5,846 \\
\midrule
Total & \textbf{588} & 9\% & 8.0 & 15.9 & 4,281 \\
\bottomrule
\end{tabular}
\end{minipage}%
}
\end{table}

The dataset is balanced across six axes: question type $\in$ {sell, buy}, aggregation $\in$ {earliest, latest}, price type $\in$ {open, close}, 13 target percentages from 0.1\% to 2.0\%, and uniformly sampled investor counts and distractor counts in the range $[2, 6]$. As a synthetic dataset, it avoids issues with model contamination; it is also designed to be updateable to the latest stock prices without having to regenerate the entire sample set. Sample instance is shown in Figure~\ref{fig:smfr_sample}, and dataset statistics in Table~\ref{tab:smfr_stats}.

Problems are generated programmatically using real historical stock prices fetched via \cite{yfinance2024} for US equities (e.g. AAPL, MSFT, GOOG, etc.), ensuring that numerical reasoning is performed on realistic distributions rather than uniform random noise. The full dataset generation pipeline from Figure~\ref{fig:stock_data_gen} is detailed below:

\begin{enumerate}
    \item \textbf{Stock Data Sampling.} For each sample, we randomly select a target transaction type (buy/sell), price type (open/close), and a target profit/loss percentage. The number of investors (parallelizable threads), the breadth $B$ (total number of stocks traded), and the depth $D$ (number of transactions per investor) of the dataset are varied to give us a range of context sizes and task difficulty. 
    
    \item \textbf{Haystack construction.} Each instance follows a "Needle-in-a-Haystack" architecture. The \textit{Haystack} consists of 30-day OHLCV histories of $B$ sampled stocks formatted as price tables, interleaved with additional distractor stocks to increase retrieval difficulty. 

    \item \textbf{Needle construction.} The \textit{Needle} consists of specific investor transaction histories embedded within the context. Each investor receives $D$ completed buy–sell pairs drawn from distinct stocks, plus one open position (the target stock) shared across all investors. The open position determines the dates on which the profit target can be achieved.

    \item \textbf{Answer computation.} The reference answer and chain-of-thought are computed deterministically from the sampled prices and transactions.   

    \item \textbf{Quality filtering.} To limit null answers, the open transaction date is sampled from the first or last 25\% of the time window. Samples with no valid qualifying dates are retried with a new seed.
\end{enumerate}

\section{Construction of Expert Designed MAS}
\label{app:manual_mas}
To establish a competitive upper bound for agentic performance on SMFR, we architect a manual MAS that utilizes structured decomposition and deterministic orchestration. Unlike the automated frameworks discussed in Section~\ref{sec:experiments}, which rely on the LLM to discover and manage its own workflow, our \textit{Expert-MAS} enforces a strict separation between linguistic processing and logical control.

\paragraph{Architecture and Role Specialization} 
Figure~\ref{fig:manual-mas} details the multi-step pipeline designed to minimize context bloat and maximize sub-task focus, composed of the following sub-agents:

\begin{enumerate}
[leftmargin=*,noitemsep,topsep=2pt]
    \item \textbf{The Meta-Agent:} A specialized agent that acts as a structural parser, responsible for extracting the problem's topology (investor names, profit targets, and aggregation criteria). This agent produces a structured JSON schema that drives the downstream orchestration, but performs no numerical reasoning itself.
    \item \textbf{The ExtractorAgent:} A reusable retrieval unit tasked with targeted information extraction from the 50k+ token haystack. It is prompted to locate specific transaction dates and prices as needed, effectively acting as a high-precision filter.
    \item \textbf{The CalculatorAgent:} A numerical reasoning unit that computes realized P\&L and derives target price thresholds. By providing this agent only with the relevant extracted snippets, we ensure its reasoning window remains uncluttered by distractor tickers.
\end{enumerate}

\paragraph{Deterministic Orchestration and Parallelism}
A significant departure from automated MAS is our use of a \textbf{Python-based Executor} for orchestration. Rather than allowing the LLM to manage the "handoff" between agents, we utilize a deterministic control script.

As shown in Figure~\ref{fig:manual-mas}, the orchestrator dispatches sub-tasks in parallel across the investor dimension. While sequential dependencies are maintained within an investor's logic chain (e.g., Transactions $\rightarrow$ P\&L $\rightarrow$ Target Price), the system executes the chains for all $N$ investors concurrently. The final comparison and win-determination are performed via deterministic Python logic.

\section{Architectural Analysis}
\label{app:analysis}

\subsection{DyLAN \citep{dylan}}
\label{app:subsec:dylan_analysis}

\begin{figure}
    \centering
    \includegraphics[width=0.95\linewidth]{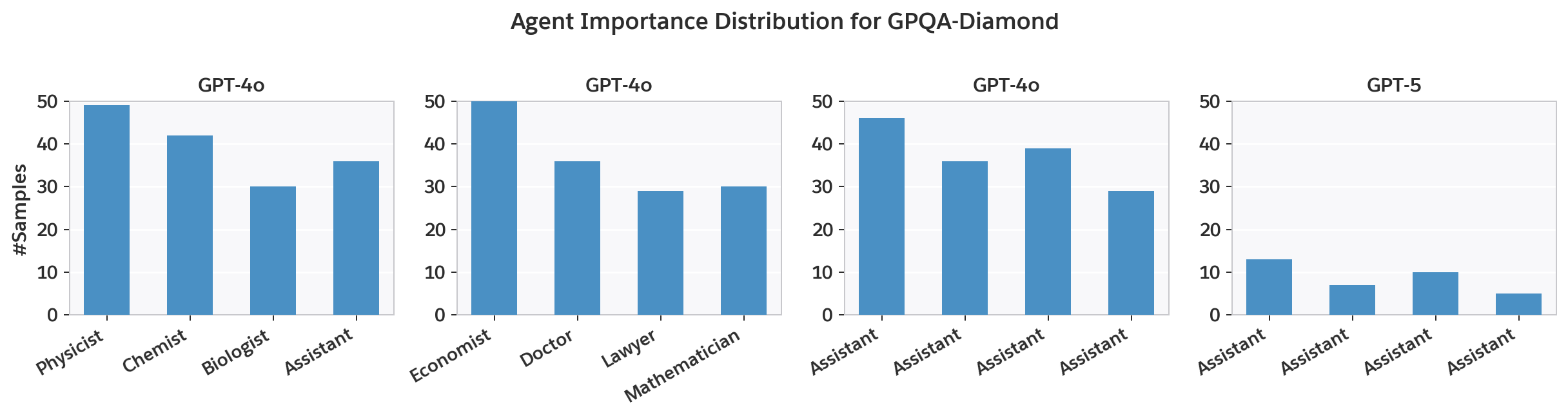}
    \caption{Highest ranked agents by importance score for different role settings in DyLAN. Results are based on  using GPT-4o and GPT-5 as backbone models for the GPQA-Diamond task. Plots reveal a slight positional bias towards the first agent regardless of role or backbone model. }
    \label{fig:dylan_roles}
\end{figure}

To investigate the causal influence of role specialization in the remaining interactive cases, we compared three configurations: (i) task-specific experts, (ii) random default roles, and (iii) generic assistant roles. Surprisingly, the all-assistant setting achieved the highest accuracy (54.41\%), outperforming task-specific experts (53.40\%). Furthermore, rankings based on agent importance scores (Figure~\ref{fig:dylan_roles}) reveal a persistent positional bias toward the first agent, regardless of assigned role or backbone model. This suggests that the reported `MAS advantage' in these paradigms is not a product of expert collaboration, but a byproduct of increased aggregate compute via redundant sampling.

\subsection{AFlow \citep{zhang2024aflowautomatingagenticworkflow}}
\label{sec:app_aflow_analysis}
\begin{figure*}[htbp]
    \centering
    % Define tcolorbox
    \begin{tcolorbox}[
        enhanced,
        % -- Color settings ---
        colframe=Salmon!90!Black,          % Border and title background color
        colback=Salmon!20,                 % Content background color
        coltitle=white,                    % Title text color
        % -- Typography and layout settings ---
        fonttitle=\large\bfseries,         % Title font
        title={The final MAS generated by AFlow on GPQA Diamond and \ourbenchmark}, % Title content
        halign title=left,                 % Left-align title
        fontupper=\footnotesize,           % Body font size inside box
        % -- Size and padding settings ---
        boxrule=1pt,                       % Border thickness
        arc=3mm,                           % Corner radius
        boxsep=2pt,
        left=8pt,                          % Left inner padding
        right=8pt,                         % Right inner padding
        top=4pt,
        bottom=4pt
    ]

\textbf{\# ==================== GPQA Diamond Workflow (GPT-5) ====================}

\hspace{0em}class Workflow:

\hspace{2em}def \_\_init\_\_(

\hspace{4em}self,

\hspace{4em}name: str,

\hspace{4em}llm\_config,

\hspace{4em}dataset: DatasetType,

\hspace{2em}) -> None:

\hspace{4em}self.name = name

\hspace{4em}self.dataset = dataset

\hspace{4em}self.llm = create\_llm\_instance(llm\_config)

\hspace{4em}self.custom = operator.Custom(self.llm)

\hspace{4em}self.sc = operator.ScEnsemble(self.llm)

\hspace{2em}async def \_\_call\_\_(self, problem: str):

\hspace{4em}"""

\hspace{4em}Implementation of the workflow

\hspace{4em}"""

\hspace{4em}solutions = [(await self.custom(input=problem, instruction=prompt\_custom.QA\_BOX\_PROMPT))['response'] for i in range(3)]

\hspace{4em}vote = await self.sc(solutions=solutions)

\hspace{4em}return vote['response'], self.llm.get\_usage\_summary()["total\_cost"]

\vspace{10pt}
\textbf{\# ==================== \ourbenchmark Workflow (GPT-4o) ====================}

\hspace{0em}class Workflow:

\hspace{2em}def \_\_init\_\_(

\hspace{4em}self,

\hspace{4em}name: str,

\hspace{4em}llm\_config,

\hspace{4em}dataset: DatasetType,

\hspace{2em}) -> None:

\hspace{4em}self.name = name

\hspace{4em}self.dataset = dataset

\hspace{4em}self.llm = create\_llm\_instance(llm\_config)

\hspace{4em}self.custom = operator.Custom(self.llm)

\hspace{4em}self.sc\_ensemble = operator.ScEnsemble(self.llm)

\hspace{2em}async def \_\_call\_\_(self, problem: str):

\hspace{4em}"""

\hspace{4em}Implementation of the workflow

\hspace{4em}"""

\hspace{4em}\# Generate multiple solutions using the custom operator

\hspace{4em}solutions = []

\hspace{4em}for \_ in range(3):  \# Generate three solutions for diversity

\hspace{6em}response = await self.custom(input=problem, instruction=prompt\_custom.XXX\_PROMPT)

\hspace{6em}solutions.append(response['response'])

\hspace{4em}\# Use ScEnsemble to select the most consistent solution

\hspace{4em}final\_solution = await self.sc\_ensemble(solutions=solutions)
        
\hspace{4em}return final\_solution['response'], self.llm.get\_usage\_summary()["total\_cost"]

    \end{tcolorbox}
    
    % Add bottom figure caption
    \caption{The final MAS workflows generated by AFlow on GPQA Diamond and \ourbenchmark, degenerating to trivial ensembling rather than sophisticated multi-agent orchestration.
    }
    \label{fig:aflow_case_analysis}
\end{figure*}

\paragraph{Degeneration to Trivial Ensembles.}

AFlow \citep{zhang2024aflowautomatingagenticworkflow} is designed to discover sophisticated workflows via tree search over graph-based code representations. However, our inspection reveals a stark divergence from this objective: instead of manifesting complex coordination, the discovered MAS consistently degenerate into trivial ensembles. As illustrated in our case analysis (Figure~\ref{fig:aflow_case_analysis}), ``optimized'' workflows frequently converge on a structure that simply iterates a single custom prompt three times before aggregation - a configuration functionally identical to standard CoT-SC. Across 14 final workflows generated by GPT-4o, GPT-5, and GPT-OSS-120B on five datasets, $50\%$ ($7/14$) adopted this simplistic structure, with four of these actually underperforming the CoT-SC baseline. 

\subsection{MAS-Zero \cite{Ke2025MASZero}}
\label{app:subsec:mas_zero}

\paragraph{Verifier Bias and Consensus Collapse (MAS-Zero).}
In MAS-Zero \citep{Ke2025MASZero}, a dedicated verifier agent aggregates outputs from parallel workers to select the optimal result. We evaluate this mechanism across BrowseComp-Plus, GPQA-Diamond, HLE-Math, and \ourbenchmark using GPT-4o and GPT-5, with selection frequencies detailed in Fig~\ref{fig:mas-zero-select}. Our analysis reveals a systematic \textit{positional bias}: the verifier disproportionately favors earlier entries in the context window, leading to premature \textit{consensus collapse}.

Across all benchmarks, we observe three consistent failure patterns:
\begin{enumerate}[leftmargin=*,noitemsep,topsep=2pt]
    \item \textbf{Extreme Primacy:} GPT-4o exhibits a severe bias toward the initial block (index 0, vanilla CoT), selecting it in over 45\% of instances, while CoT-SC (index 1) remains a distant secondary choice.
    \item \textbf{Broadened Initial Bias:} GPT-5 demonstrates a slightly more distributed but still front-loaded preference, favoring the first four fundamental reasoning blocks (indices 0--3) while largely ignoring subsequent iterations.
    \item Blocks corresponding to later search rounds (indices 4--8) are rarely selected by either model, accounting for less than 15\% of total selections combined.
\end{enumerate}

Consequently, the complex MAS architecture suffers from structural redundancy: subsequent worker agents function as "expensive witnesses", incurring full inference costs while exerting zero causal influence on the final output.

\subsection{ADAS \citep{hu2024ADAS}}
\label{app:subsec:adas_analysis}

ADAS \citep{hu2024ADAS} optimizes MAS architectures through consecutive iterations of agent discovery. While the framework is designed to iteratively refine performance, our analysis on GPQA-Diamond reveals that the search results are non-monotonic, lacking a consistent trajectory of improvement. As illustrated in Figure~\ref{fig:adas-gpqa-search-dynamics}, validation accuracy frequently peaks early in the search phase before regressing or plateauing, rather than accumulating incremental gains.

This differs from the pattern reported on the ARC dataset in the original work, where stronger-performing agents were gradually discovered in later iterations. We hypothesize this discrepancy stems from a potential evaluation artifact: through correspondence with the authors, we confirmed that their reported results were derived from evaluating all generated MAS candidates directly on the test set and selecting the global maximum. Consequently, our findings suggest that ADAS functions primarily as a heuristic explorer of architectural variants rather than a reliable optimizer, where performance gains are susceptible to ``lucky'' iterations rather than structural evolution.

\paragraph{Non-monotonic Search Across Iterations.} ADAS \citep{hu2024ADAS} aims to iteratively refine MAS architectures through automated agent discovery. However, our analysis on GPQA-Diamond reveals that architectural search is non-monotonic: validation accuracy frequently peaks early and subsequently regresses or plateaus, rather than accumulating incremental gains (see Figure~\ref{fig:adas-gpqa-search-dynamics}). This deviates from the original work’s reported performance on the ARC dataset, which we hypothesize is an artifact of selecting the global maximum from the test set across all candidates.\footnote{Through correspondence with the authors, we confirmed their results were derived by evaluating all candidates directly on the test set.} These findings suggest that ADAS functions as a heuristic explorer rather than a reliable optimizer; performance gains appear to be the result of stochastic ``lucky'' iterations rather than a principled structural evolution toward superior reasoning.

\paragraph{Architectural Redundancy.} To isolate the structural drivers of performance, we conducted a motif analysis by mapping generated architectures to a rule-based dictionary (e.g., Self-consistency, Aggregation, Verifier). Across all settings, the primary positive signal originated from Self-consistency motifs. On GPQA-Diamond, architectures incorporating these motifs achieved a mean accuracy of $82.19\%$ ($+1.34\%$ over the global average), whereas ``specialized'' coordination motifs yielded negligible gains. This mechanistic evidence confirms that automated search often converges on rediscovering CoT-SC style sampling under more complex labels, rather than inventing novel or synergistic multi-agent strategies.

\subsection{MaAS \citep{zhang2025agentic-supernet}}
\label{app:subsec:maas_analysis}

\paragraph{Incentive Misalignment and Routing Collapse.} MaAS \citep{zhang2025agentic-supernet} optimizes its controller via Monte Carlo gradient estimation, balancing an accuracy objective against a cost penalty. However, we find that with highly capable base models (e.g., GPT-5), accuracy frequently saturates, flattening the gradient to $\sim 1/K$ and extinguishing the signal required to learn task-specific routing. Consequently, the controller’s behavior is dictated almost entirely by the cost term, resulting in two distinct failure modes: (1) \textbf{Cost-Minimizing Collapse} on BrowseComp-Plus, where high cost variance drives the controller toward a trivial single I/O call ($74.2\%$ of activations); and (2) \textbf{Stochastic Stalling} on GPQA-Diamond, where negligible cost differentials trap the controller in its initialized near-uniform distribution. In both cases, the supernet fails to acquire meaningful routing logic, settling into either a ``cheap shortcut'' or an undifferentiated ensemble that consistently underperforms independent CoT-SC sampling (Table~\ref{tab:maas_operator_activation}).

\subsection{MAS-Orchestra \citep{Ke2026MASOrchestra}}
\label{app:subsec:masorchestra_analysis}
\paragraph{Policy Collapse in Dynamic Orchestration.}

MAS-Orchestra \citep{Ke2026MASOrchestra} is designed to perform dynamic resource allocation by routing queries to agents based on task difficulty. However, our analysis reveals a total policy collapse into difficulty-agnostic behavior. Across all benchmarks, the system largely ignores its diverse agent pool, converging instead on a rigid binary preference for high-overhead Debate and Reflexion agents (see Table~\ref{tab:masorchestra_agent_selection}). Crucially, the orchestrator fails to scale agent complexity to task difficulty; despite GPQA-Diamond posing a lower reasoning ceiling than HLE-Math, the system exhibited a higher reliance on Debate agents for the former ($84.9\%$) than the latter ($79.2\%$). These results demonstrate that the orchestrator does not manifest adaptive configuration; instead of learning task-specific strategies, it settles into a static, greedy preference for maximum-overhead sub-agents regardless of a query's actual requirements.

\section{Scope and Limitations}
\label{sec:limitations}

\paragraph{Model Diversity and Selection Bias.} Our study primarily utilizes frontier models from the OpenAI and Google families, alongside a single representative open-source backbone. While this selection spans varying scales and generations, it is possible that specific architectural idiosyncrasies of other model families (e.g., Anthropic’s Claude or Meta’s Llama series) might yield different interaction dynamics. Furthermore, because cost-efficiency was a central pillar of our evaluation, we did not explore infinite-budget regimes where extremely large ensembles might eventually overcome the identified positional biases through sheer scale.

\paragraph{Reasoning vs. Tool-Use Proficiency.} Our evaluation focuses primarily on cognitive orchestration and long-horizon reasoning within closed or semi-closed contexts. While benchmarks like BrowseComp-Plus and SWE-bench Lite involve retrieval and patch generation, we did not evaluate the broader spectrum of autonomous tool-use, such as real-time API interaction, multi-modal sensor integration, or complex shell environments. It remains possible that the structural efficiencies identified in our "Expert-MAS" might differ in environments where the primary bottleneck is external tool-call latency or protocol adherence rather than internal logical consistency. Our findings of functional collapse are therefore most applicable to reasoning-heavy agentic workflows.

\paragraph{Optimization Hyperparameters.} Our evaluation of automated frameworks (e.g., ADAS, AFlow) utilized the default search hyperparameters provided by the original authors. It is conceivable that with extensive, domain-specific hyperparameter tuning, these frameworks could find more robust coordination motifs. However, we intentionally maintain default configurations across all systems - including CoT-SC and Expert-MAS - to evaluate out-of-the-box reliability. Our findings suggest that while expert-designed and simple SAS baselines remain robust under default settings, current automated MAS search processes are highly sensitive, failing to consistently outperform SAS without extensive optimization.

\section{Broader Impacts}
\label{app:impacts}

This work introduces a diagnostic benchmark designed to evaluate the reasoning efficiency of multi-agent systems. While the dataset utilizes financial market primitives, it is intended strictly for AI safety and architectural research and is not validated for real-world financial forecasting or automated trading. By identifying structural bloat in AI workflows, this research promotes the development of more computationally efficient and transparent models, potentially reducing the environmental and economic costs of large-scale AI deployment. We do not foresee any significant negative societal impacts, provided the benchmark is used as a diagnostic tool rather than a predictive model for safety-critical domains.

%%%%%%%%%%%%%%%%%%%%%%%%%%%%%%%%%%%%%%%%%%%%%%%%%%%%%%%%%%%%

\end{document}